\documentclass[11pt]{article}

\usepackage[normalem]{ulem}
\usepackage{xcolor}
\usepackage{booktabs}
\usepackage{pifont}
\usepackage[most]{tcolorbox}
\usepackage{listings}
\lstdefinestyle{prompt}{
  basicstyle=\ttfamily\small,
  columns=fullflexible,
  breaklines=true,
  breakatwhitespace=true,
  keepspaces=true,
  showstringspaces=false
}
\newtcblisting{promptbox}{
  listing only,
  colback=gray!4,
  colframe=gray!50,
  arc=2mm,
  boxrule=0.5pt,
  left=2mm,right=2mm,top=1mm,bottom=1mm,
  listing options={style=prompt}
}
\usepackage{hyperref}
\usepackage[final]{acl}

\usepackage{times}
\usepackage{latexsym}

\usepackage[T1]{fontenc}

\usepackage[utf8]{inputenc}

\usepackage{microtype}

\usepackage{inconsolata}

\usepackage{graphicx}
\usepackage{wrapfig}

\usepackage{enumitem}

\usepackage{adjustbox}

\usepackage{booktabs,longtable,array,ragged2e}
\usepackage{pdflscape}
\usepackage{xspace}

\newcommand{\ourbench}{\textsc{FDARxBench}}

\title{FDARxBench: Benchmarking Regulatory and Clinical Reasoning on FDA Generic Drug Assessment}

\author{
 \textbf{Betty Xiong\textsuperscript{1}},
 \textbf{Jillian Fisher\textsuperscript{2}},
 \textbf{Benjamin Newman\textsuperscript{2}},
\\
 \textbf{Meng Hu\textsuperscript{3}},
 \textbf{Shivangi Gupta\textsuperscript{3}},
\\
 \textbf{Yejin Choi\textsuperscript{1}},
 \textbf{Lanyan Fang\textsuperscript{3}},
 \textbf{Russ Altman\textsuperscript{1}}
\\
\\
 \textsuperscript{1}Stanford University,
 \textsuperscript{2}University of Washington,
 \textsuperscript{3}U.S. Food and Drug Administration
\\
 \small{
   \textbf{Correspondence:} \href{mailto:xiongb@stanford.edu}{xiongb@stanford.edu}, \href{mailto:Russ.Altman@stanford.edu}{Russ.Altman@stanford.edu}
 }
}

\begin{document}
\maketitle
\begin{abstract}
We introduce an expert-guided, real-world benchmark for evaluating document-grounded question-answering (QA) motivated by generic drug assessment, using the U.S. Food and Drug Administration (FDA) drug label documents. Drug labels contain rich but heterogeneous clinical and regulatory information, making accurate question answering difficult for current language models. In collaboration with FDA regulatory assessors, we introduce \ourbench{} \footnote{\ourbench{} may be accessed via: https://github.com/xiongbetty/FDARxBench (DOI 10.5281/zenodo.22073747)}, and construct a multi-stage pipeline for generating high-quality, expert-guided, LLM-assisted, QA examples spanning factual, multi-hop, and refusal tasks, and design evaluation protocols to assess both open-book and closed-book reasoning. Experiments across proprietary and open-weight models reveal substantial gaps in factual grounding, long-context retrieval, and safe refusal behavior. While motivated by FDA generic drug assessment needs, this benchmark also provides a substantial foundation for challenging regulatory-grade evaluation of label comprehension.

\begin{figure}[t]
\centering
\includegraphics[width=\linewidth]{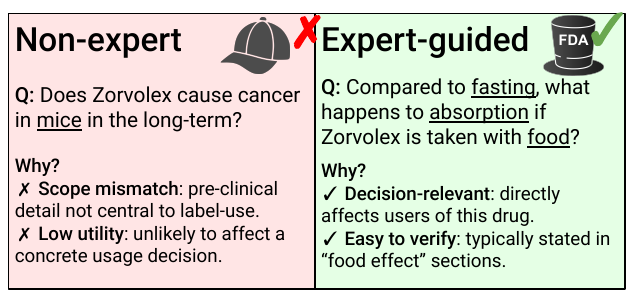}

\caption{Expert-guided question relevance for FDA drug-label QA. FDA drug labels contain heterogeneous clinical and regulatory information. Expert reviewers help distinguish questions that are merely label-adjacent from questions that are useful for regulatory QA. The rejected non-expert example asks about preclinical carcinogenicity in a way that is less central to the intended label-use setting, while the retained expert-guided example asks about food effects on absorption, a decision-relevant question grounded in pharmacokinetic labeling.}
\label{fig:expert_guided_relevance}
\end{figure} \end{abstract}
\section{Introduction}

Large language models (LLMs) have demonstrated strong performance on biomedical question answering tasks \cite{biogpt, meditron, medhelm, medgemma}, but their abilities in high-stakes regulatory settings remain under-explored.
Their performance and practical utility are still not clearly defined, in part because progress is bottlenecked by the lack of standardized, regulator-aligned benchmarks and limited access to experts who can define regulator-grade correctness, provenance, and safe abstention.

The motivating use case for this work is the US Food and Drug Administration (FDA) generic drug assessment: a real-world high-stakes document QA setting where FDA regulators evaluate generic drug applications by comparing them to approved reference drugs. This requires analyzing detailed, multi-page drug labels to answer specific high-stakes questions. The process is time intensive and demands careful review, making it both a natural and challenging candidate for AI assistance.

Prior work has demonstrated promising applications of GPT-style models to support drug label analysis \cite{psg_gpt}, but their performance assessment in that study relied primarily on manual review, motivating the need for systematic performance benchmarks. To overcome this, we partner with professional FDA regulatory assessors with experience in generic drug assessment to shape a benchmark and expert-adjudicate question quality and model outputs.
Distinct from other biomedical settings, regulatory assessors (1) care about complex relationships between specific entities and quantities, (2) need answers with fine-grained provenance, and (3) require reliable abstentions to unanswerable questions.
However, building large-scale benchmarks to assess these aspects is difficult as FDA regulatory expertise is not widespread.

In this work, we propose a benchmark to systematically evaluate LLM performance for regulator-grade question-answering grounded in drug labels (Figure~\ref{fig:expert_guided_relevance}).
Expert input from FDA regulatory assessors informs many aspects of our benchmark: they provide criteria for regulatory relevance, feedback on benchmark design, and validation of the LLM screening process.
Our benchmark consists of 17K+ questions that (1) focus on entities and quantities that are clinically meaningful to FDA assessors in generic drug assessment workflows, (2) require cross-section reasoning with fine-grained provenance, and (3) require clear abstention to unanswerable questions.
This benchmark is LLM-assisted in generation, and validated against expert annotation on a smaller QA subset used for calibration. It enables evaluation of reasoning, retrieval, long-context comprehension, grounding, and safety behavior within a unified framework.
The proposed pipeline is modular and can be adapted or extended to other specialized domains where expert-defined correctness and provenance are required.

Our contributions are threefold: (1) a 17K+ regulatory QA benchmark grounded in FDA drug labels; (2) task definitions and evaluation protocols that emphasize correctness, provenance, and safe abstention over surface-level fluency; and (3) detailed human expert and model-based analyses that reveal persistent failures in grounding and safe abstention.

\section{Background and Related Work}

\subsection{FDA Drug Labels}

FDA drug label documents are the legally authoritative documents on prescription drugs in the United States. They are formatted as Structured Product Labeling (SPL) files \cite{spl} -- a kind of semi-structured XML -- under the Physician Labeling Rule (PLR). Their sections can vary widely across manufacturers and time periods, yielding multi-page, heterogeneous documents that challenge automated reasoning. While prior work has incorporated LLMs into label-centric workflows \cite{psg_gpt} and evaluation settings \cite{unitox}, no benchmark specifically measures LLM performance on drug label understanding.
Past work has identified regulatory priorities review-centric functionality, transparency and explainability, and data security \citep{askfdalabel}. \ourbench{} addresses these priorities by focusing on drug-label questions motivated by regulatory review workflows, requiring explicit passage-level citations for provenance and verifiability, evaluating safe refusal when label evidence is absent, and using public FDA label documents rather than private patient data. 

\subsection{Biomedical QA Datasets}

Current biomedical benchmarks focus on domains outside of regulation.
For example \textsc{BioASQ} and \textsc{PubMedQA} target questions derived from biomedical literature \cite{bioasq, pubmedqa}.
\textsc{HealthSearchQA} targets general healthcare consumers \cite{Singhal2022HealthSearchQA}.
More recent evaluation efforts such as \textsc{HealthBench} and \textsc{MedHELM} broaden biomedical QA evaluation across clinical, medical reasoning, and healthcare communication settings \citep{Arora2025HealthBenchEL, medhelm}.
Still other tasks are more clinical-facing, such as QA over electronic medical records and laboratory tests \citep{Pampari2018emrQAAL, Bhasuran2025LabQARAM}; and multimodal reasoning benchmarks such as MicroVQA that rely on expert-guided annotations \citep{burgess2025microvqa}.
However, unlike existing biomedical benchmarks, evaluations of FDA drug labels sit at the intersection of medicine and regulation, and current benchmarks do not capture key regulatory requirements such as reasoning over full-length documents, providing traceable citations, and explicitly refusing unanswerable questions. To our knowledge, no prior benchmark directly targets real-world regulatory drug QA with expert input, which is the primary gap our task is designed to address.

\section{Dataset: Expert-Guided FDA Label QA Benchmark}
\label{sec:dataset}
The benchmark is designed to support evaluation of LLM behavior on drug-label questions motivated by generic drug assessment (dataset statistics in Table~\ref{tab:all_dataset_stats}). In this section, we outline how we create \ourbench{} (Figure~\ref{fig:schematic}). (More details in Appendix~\ref{appendix:dataset}).

\begin{figure*}[t]
    \centering
    \includegraphics[width=0.8\linewidth]{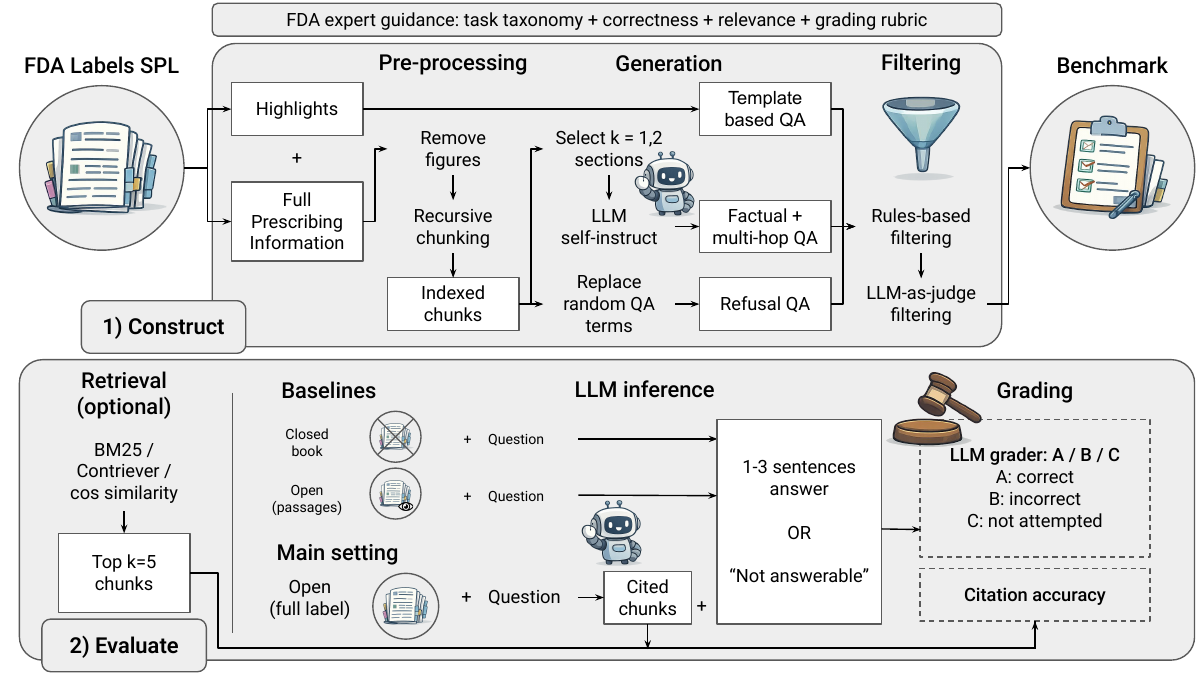}
    \caption{Overview of \ourbench{} creation.}
    \label{fig:schematic}
\end{figure*}

\begin{table}[t]
\centering
\tiny
\setlength{\tabcolsep}{2pt}
\renewcommand{\arraystretch}{1.15}
\begin{tabular}{l r|lrr}
\toprule
\textbf{Dataset statistic} & \textbf{Count}&  \textbf{QA statistic}&\textbf{Mean}&\textbf{Std}\\
\midrule
SPL drug labels & 700  &  No. of tokens (question)&20.35&6.85\\
Final QAs & 17{,}223  &  No. tokens (answer)&39.23&52.69\\
Factual & 9{,}888  &  No. required context passages (question)&1.77&3.85\\
Multi-hop & 3{,}400  &  No. tokens (drug label)&7661.77&7542.68\\
Refusal & 3{,}935  &  No. passages (drug label)&45.80&37.95\\
\bottomrule
\end{tabular}
\caption{Dataset summary statistics.}
\label{tab:all_dataset_stats}
\end{table}

\subsection{Source Documents and Preprocessing}
We curate 700 FDA prescription drug labels, sourced from the FDALabel Database \cite{fdalabeldb}, as the starting corpus of our benchmark. In general, drug labels are semi-structured XML files that contain text details on new drugs. We preprocess each drug label by parsing it into section-level passages using subheaders as delimiters, where every passage is assigned a unique identifier. This new chunked representation allows us to (1) track provenance and evaluate citations, and (2) enable retrieval-augmented workflows.

\subsection{Question Generation Pipeline}
\label{sec:pipeline}
After preprocessing the data, we use established synthetic and self-instruct data generation approaches \citep{wang2023selfinstructaligninglanguagemodels} to construct our benchmark using a modular four-step pipeline:

\begin{enumerate}[leftmargin=1em,itemsep=0em,parsep=0em,topsep=0em]
    \item \textbf{Context selection.} We randomly select a portion of the drug label to be used as the main context for question generation.
    \item \textbf{LLM  question generation.} Given the context chosen in step (1), we use an LLM with few-shot prompting to generate a QA pair. The context acts as supporting evidence that must be used to answer the generated question. This generation changes based on question types, shown in \S~\ref{sec:types}.
    \item \textbf{Expert annotation protocol.} Three experienced FDA reviewers rate 50 LLM-generated questions as yes/maybe/no to indicate whether it should be kept. They are provided with the LLM filtering prompt as instructions. (See Appendices~\ref{app:eval_details} and \ref{app:rubric} for filtering instructions and rubric).
    \item \textbf{LLM-based filtering for experts.} The same prompt is used for scalable filtering. Human expert relevance agreement (collapsed to binary yes+maybe / no) is substantial. The mean pairwise percent agreement (0.88) and inter-rater agreement (Fleiss’ $\kappa$ = 0.72) indicate that experts tended to have agreement on acceptable questions for the benchmark. We audit the GPT validator against the expert-majority decision, with random vote if no majority: GPT-retained questions tend to match human ones (0.85 F1, 0.76 PPV, 0.97 Recall).  (See Appendix~\ref{app:eval_details} for inter-annotator statistics).
    \item \textbf{Filtering.} Finally, we apply rule-based and LLM-as-a-judge filtering. We use automatic rule-based filters to remove QAs that are structurally invalid, i.e., items with a missing question or answer, empty strings, absence of the drug name are discarded. For the LLM-as-a-judge, we filter based on QA correctness (answer is correct given only the provided context) and question quality (remove QA pairs whose question is semantically invalid; and filter multi-hop questions to ensure neither passage alone is sufficient to answer it, and that both are necessary). (Further details on the filtering steps in Appendix~\ref{appendix:filtering}). We validate the LLM filter against human annotators, and report an average precision of 0.80 and average F1 of 0.68. However, this may be low due to multi-hop questions, which are known to be a harder task; the metrics increase to 0.92 and 0.74, respectively, without multi-hop questions. (More details in Appendix~\ref{app:eval_details}).
\end{enumerate}

\subsection{Question Types}
\label{sec:types}
Using this general pipeline, we construct three diverse QA types (factual, multi-hop and refusal), each targeting a distinct regulatory capability relevant to generic drug assessment.
(For examples, see Appendix~\ref{appendix:question_types}).

\textbf{Factual (one-section) questions} are fact-based questions that can be answered directly using information from a single section of the drug label. To construct them, we select either the Highlights section, which provides a structured summary of key information, or another section sampled from the full label, and explicitly instruct the LLM to generate fact-based questions grounded only in the content of the selected section, as outlined in \S~\ref{sec:pipeline}. 

\textbf{Multi-hop (cross-section) questions} require integrating information from two different sections of the same drug label by design, motivated by prior multi-hop question generation work, e.g., HotPotQA, MuSiQue, MQuAKE \citep{yang-etal-2018-hotpotqa, trivedi2023interleavingretrievalchainofthoughtreasoning, zhong2024mquakeassessingknowledgeediting}. To construct them, we randomly select pairs of sections and explicitly instruct the LLM to generate questions that depend on evidence from both sections, such that the question becomes unanswerable if either section is removed.

\textbf{Refusal (unanswerable) questions} are negative controls designed to be unanswerable: where the requested information is absent from the FDA label, abstention is the correct behavior. %
We generate them by inserting an out-of-scope biomedical entity/keyword into a clinical-style template and programmatically verifying it does not appear anywhere in the full label text (case-insensitive string match), so the correct behavior is to abstain rather than hallucinate or rely on outside knowledge.

\subsection{Tasks}
\label{sec:tasks_eval}
We showcase the versatility of \ourbench{} by evaluating models under multiple diverse evidence settings. We describe these settings below:
\begin{itemize}[leftmargin=0.5em,itemsep=0em,parsep=0em,topsep=0em]
    \item \textbf{Full-label QA with citations.} The entire label is provided as context, to evaluate the real-world generic drug assessment setting of model output plus cited passage ids.
    \item \textbf{Closed-book QA.} Models answer from the question alone, without label text, external context or retrieval, as a lower-bound baseline. This parameter-memory ablation setting aligns with prior QA work of \citet{roberts-etal-2020-much}, where it is a contamination probe to measure whether models answer from memorized label knowledge or hallucination when relevant FDA label evidence is withheld.
    \item \textbf{Oracle passages.} Gold passage(s) used during construction are provided as an upper-bound diagnostic.
    \item \textbf{Retrieval only evaluation.} The model is evaluated on cited passages only (e.g., a retriever selects top-$k$ passages, which are compared against the gold standard citations), to evaluate the potential performance of RAG methods.
\end{itemize}

\subsection{Metrics}
\label{sec:metrics}
For evaluation of \ourbench{}, we report metrics %
aligned with regulatory priorities identified in FDA-authored work on regulatory QA systems \citep{askfdalabel} and operationalized through feedback from FDA regulatory reviewers during task design and relevance review:
\begin{itemize}[leftmargin=1em,itemsep=0em,parsep=0em,topsep=0pt]
    \item \textbf{Answer correctness.} Similar to past work \cite{simpleqa}, accuracy is evaluated by LLM-as-judge given the benchmark's reference answer.
    (See Appendix~\ref{app:eval_details} for human validation details). 
    \item \textbf{Refusal behavior.} Evaluation using macro-F1 for predicting abstention on refusal items, and false-refusal behavior on answerable items.
    \item \textbf{Citation quality.} Automatic evaluation using macro-F1 between cited passage ids and gold provenance passage ids. %
\end{itemize}

\section{Experiments and Results}
\label{sec:results}
In this section, we evaluate \ourbench{} on a variety of state-of-the-art (SOTA) LLM models. Given this analysis, we find that \ourbench{} is challenging, even for the strongest of current models.
The open weight and API-based models that we evaluate are in Table \ref{tab:results_by_evidence_setting}.

\begin{table}[h]
\centering
\tiny
\setlength{\tabcolsep}{2pt}
\begin{tabular}{lccp{3em}p{3em}c|cc|cc}
\toprule
\textbf{Model} &   \multicolumn{5}{c|}{\textbf{Full label (+ cites)}}&\multicolumn{2}{c|}{\textbf{Closed-book Acc}} & \multicolumn{2}{c}{\textbf{Oracle Acc}}  \\
&   \textbf{Fact} &  \textbf{MH} & \textbf{Cite F1 Fact}&\textbf{Cite F1 MH}&\textbf{Ref F1} &\textbf{Fact} & \textbf{MH} & \textbf{Fact} & \textbf{MH} \\
\midrule
Llama-8B &   0.500 &  0.256 & 0.458&0.333&\textbf{0.796} &0.002 & 0.003 & 0.777 & 0.668 \\
Llama-70B &   0.526 &  0.422 & 0.508&0.433&0.789 &0.237 & 0.393 & 0.797 & 0.680 \\
Mistral-14B &   0.503 &  0.341 & 0.439&0.328&0.710 &0.087 & 0.150 & 0.776 & 0.676 \\
Qwen-14B &   0.518 &  0.416 & 0.500&0.362&0.709 &0.195 & 0.385 & 0.795 & 0.789 \\
Qwen-32B &   0.530 &  \textbf{0.473} & 0.525&0.433&0.756 &0.186 & 0.370 & 0.786 & 0.796 \\
Claude Sonnet &   0.526 &  0.372 & \textbf{0.528}&0.383&0.748 &0.259 & 0.272 & 0.784 & 0.653 \\
 Claude Opus&   \textbf{0.562}&  0.427& 0.522&\textbf{0.458}&0.731&0.369& 0.393& 0.794& 0.728\\
GPT-4o-mini &   0.507 &  0.456 & 0.497&0.378&0.717 &0.242 & 0.442 & 0.806 & \textbf{0.822} \\
GPT-5.1 &   0.546&  0.461 & 0.520&0.406&0.725 &\textbf{0.343} & \textbf{0.515} & 0.810& 0.805 \\
 GPT-5.2&   0.541&  0.417& 0.520&0.383&0.701&0.292& 0.452& \textbf{0.817}& 0.806\\
\bottomrule
\end{tabular}
\caption{Model performance across evidence settings. We report answer accuracy (Acc) for factual (Fact) and multi-hop (MH) questions in closed-book, oracle-passages, and full-label settings; the full-label setting additionally reports citation overlap (Cite F1) and refusal correctness (Ref F1).}
\label{tab:results_by_evidence_setting}
\end{table}

\begin{table}[h]
\centering
\tiny
\setlength{\tabcolsep}{2pt}
\renewcommand{\arraystretch}{1.15}
\begin{tabular}{lcc|cc|cc|cc}
\toprule
\textbf{Retriever} &
\multicolumn{2}{c|}{\textbf{recall@1}} &
\multicolumn{2}{c|}{\textbf{recall@5}} &
\multicolumn{2}{c|}{\textbf{recall@10}} &
\multicolumn{2}{c}{\textbf{recall@\textbar gold\textbar}} \\
\cmidrule(lr){2-3}\cmidrule(lr){4-5}\cmidrule(lr){6-7}\cmidrule(lr){8-9}
& \textbf{Fact} & \textbf{MH} & \textbf{Fact} & \textbf{MH} & \textbf{Fact} & \textbf{MH} & \textbf{Fact} & \textbf{MH} \\
\midrule
BM25 & \textbf{0.558}& \textbf{0.355}& \textbf{0.748}& \textbf{0.778}& \textbf{0.797}& \textbf{0.883}& \textbf{0.592}& \textbf{0.546}\\
MiniLM-L6-v2 & 0.423& 0.294& 0.676& 0.668& 0.750& 0.792& 0.463& 0.450\\
ReContriever & 0.220& 0.176& 0.455& 0.438& 0.589& 0.593& 0.238 & 0.278 \\
\midrule
Hybrid (dense/lexical)\\
 Hybrid (0.75/0.25)& 0.567& 0.362& 0.752& 0.774& 0.804& 0.879& 0.599&0.558\\
  Hybrid (0.50/0.50)& \textbf{0.580}& \textbf{0.371}& \textbf{0.760}& \textbf{0.802}& \textbf{0.808}& \textbf{0.898}& \textbf{0.614}&\textbf{0.579}\\
 Hybrid (0.25/0.75)& 0.571& 0.366& 0.755& 0.796& 0.803& 0.895& 0.604&0.565\\
 Hybrid (0.10/0.90)& 0.565& 0.360& 0.750& 0.785& 0.801& 0.888& 0.598&0.555\\
 RRF& 0.545& 0.346& 0.740& 0.766& 0.799& 0.880& 0.579&0.545\\
\bottomrule
\end{tabular}
\caption{Retriever performance on evidence selection for factual (Fact) and multi-hop (MH) questions. We report recall at ranks $k \in \{1,5,10\}$ and at the gold set size ($\lvert gold\rvert$). RRF refers to Reciprocal Rank Fusion.}
\label{tab:retriever_recall_by_k}
\end{table}

\subsection{Model Results}
\label{sec:main_results}

\textbf{Current LLMs consistently struggle with \ourbench{}, with evidence access driving performance}. Table~\ref{tab:results_by_evidence_setting} summarizes performance across \ourbench{} tasks. In the closed-book setting, accuracy is generally low (average = 0.22 for factual, 0.34 for multi-hop), especially for smaller open-weight models, indicating that SOTA models do not have access to the drug labels in parametric knowledge. Oracle passages substantially improve performance across factual and multi-hop questions ($\approx$ 0.78-0.82 factual and 0.65-0.82 multi-hop), but scores remain below those reported on established biomedical QA benchmarks such as \textsc{BLURB} \citep{Gu_2021}, \textsc{MedQA} \citep{jin2020diseasedoespatienthave}, \textsc{PubMedQA} \citep{pubmedqa}, with best scores of 0.84, 0.97, and 0.82, respectively. Full-label access does not close this gap: despite containing strictly more information, full-label accuracy reaches only 0.53 for factual and 0.40 for multi-hop on average (best full-label multi-hop = 0.47). The oracle to full-label drop indicates that performance is limited due to the challenge of long-context tasks. (Error analyses and comparison to reasoning benchmarks in Appendices~\ref{app:error_analysis} and \ref{app:benchmark_comparison}).

\textbf{Provenance and refusal behavior remain inconsistent even when answers improve.} Citation overlap is moderate even for the best models, (e.g., cite F1 peaks at 0.53 for factual and 0.46 for multi-hop), indicating that models often cite plausible but non-gold or incomplete passages. Refusal behavior also varies substantially across models and does not simply track answer accuracy, e.g., refusal F1 ranges from 0.71 for Qwen-14B to 0.80 for Llama-8B, with the smallest 8B model giving the best F1 score. Refusal behavior exhibits a clear precision-recall trade-off, while frontier models achieve high refusal recall ($\approx$ 0.99) but over-refuse. This suggests that refusal should be evaluated as a distinct safety capability, rather than assumed to follow from general QA quality. (We report false-refusal rates separately for factual and multi-hop questions in Appendix~\ref{app:error_analysis}). %

\subsection{Retriever Results}
\label{sec:retriever_results}

\textbf{Retrieval is a bottleneck to accurate fine-grained citations.} To isolate evidence selection from generation, Table~\ref{tab:retriever_recall_by_k} compares the retrieved top-$k$ passages (e.g., $k=1,5,10, \lvert gold\rvert$) against gold provenance. We evaluate on: BM25 \cite{bm25}, MiniLM-L6-v2 cosine similarity \cite{all-MiniLM-L6-v2}, ReContriever \cite{recontriever}, linear hybrid BM25+dense scoring, and reciprocal rank fusion (RRF). 
Hybrid retrieval linearly combines z-score-normalized BM25 and dense scores (MiniLM-L6-v2) with different lexical/dense weights (0.5/0.5, 0.75/0.25, 0.25/0.75).
We report retrieval quality using F1 calculated between citation overlap against gold provenance.

Across cutoffs in single retrieval methods, BM25 outperforms dense baselines for both factual and multi-hop retrieval (e.g., recall@1 = 0.56 factual / 0.36 multi-hop vs.\ MiniLM-L6-v2 at 0.42 / 0.29 and ReContriever at 0.22 / 0.18), suggesting that drug-label evidence selection is strongly driven by lexical overlap and section-style phrasing rather than embedding similarity. Interestingly, hybrid approaches outperform the single method approach, with 0.5/0.5 dense/lexical weighting achieving the highest recall@1 (0.58 factual / 0.37 multi-hop), while RRF performs slightly worse than BM25 alone. These retrieval trends indicate that multiple sources of information might be helpful for this task.
Overall, these results mirror the end-to-end gap between full-label and oracle-passages in the Model results (\S~\ref{sec:main_results}).

\section{Conclusion}

We introduce an expert-guided benchmark for regulatory-grade QA motivated by FDA generic drug assessment, grounded in FDA drug labels, with 17K+ document-grounded questions spanning factual, multi-hop, and refusal types and tasks covering closed-book, citation-grounded open-book, retrieval-augmented QA, and safe refusal. Results highlight a gap between general biomedical QA and regulatory needs, where correctness, provenance, and conservative abstention are essential.

\section*{Limitations}

Our benchmark has several important limitations.
First, self-instruct questions are generated from individual section chunks, which can artificially narrow context and does not guarantee unique evidence localization: labels are often redundant, restating key facts across sections, so some questions admit multiple valid evidence locations even though gold provenance is defined at the chunk level. Future work could generate questions over larger windows or the full label, and represent provenance as equivalence classes of acceptable supporting spans rather than a single chunk, at higher curation cost.

Second, our LLM-based grading can be imperfect, particularly for borderline cases, and may be subject to a ceiling effect from reference-based scoring: because reference answers are generated from narrow passage selections, a model that answers more comprehensively than the reference, using accurate and citation-grounded evidence, could be penalized as incorrect. Future work should incorporate calibrated human adjudication for ambiguous items and stronger faithfulness checks, such as entailment-based verification that cited text supports each claim, to help disentangle genuine errors from reference-completeness artifacts.

Third, multi-hop construction remains challenging: despite prompting for cross-section reasoning, many candidate multi-hop questions are filtered for invalid hop structure, which may underrepresent realistic multi-constraint regulatory reasoning. More effective multi-hop items likely require deeper expert involvement and may link to more than two sections, (e.g., one may ask a question related to food effect on pharmacokinetics, efficacy and safety or whether the dosage needs to be adjusted under fed condition).

Fourth, refusal questions provide a controlled test of hallucination resistance but rely on heuristic templates and keyword exclusion, which may not capture the full diversity of unanswerable queries in practice. Expanding refusal evaluation to include naturally occurring ambiguous or partially answerable questions is an important direction.

Finally, as with all benchmarks, there is a tradeoff between releasing the benchmark and increasing future contamination risk. Regulatory updates and drug label withdrawals are important for maintaining a benchmark grounded in living FDA documents. We release the current benchmark as a frozen snapshot with label-level metadata, including drug identifiers, set IDs, and source-label information so future versions can track label revisions or withdrawn/updated documents. If the dataset were to be expanded to new labels, refreshed releases can be versioned separately with our released question-generation prompts and filtering rubric in Appendix~\ref{app:prompts}.

\section*{Ethical Considerations}

\ourbench{} is designed to be used in beneficial ways like increasing efficiency and quality of the drug approval process. Its goal is to support evaluation of language models for document-grounded regulatory question answering, not to provide medical advice or replace expert regulatory, clinical, or legal judgment. The benchmark uses publicly available FDA drug-label documents and does not contain private patient data. Any deployment of models evaluated on this benchmark should include expert human oversight and careful validation. 
\section*{Acknowledgments}
We thank Liang Zhao for his expertise in annotating FDA question relevance rubrics for expert-guided question generation. We thank David Hall, Hanwen Xu, Nelson Liu, Percy Liang and Sheng Wang for useful conversations.

Funding:  BX is supported by Australian American Fulbright Commission Future Scholarship.  RBA is supported by Burroughs Wellcome Fund Grant 1074128.  RBA is supported by NIH GM153195.  This project was also supported by the Food and Drug Administration (FDA) of the U.S. Department of Health and Human Services (HHS) as part of a financial assistance award Center of Excellence in Regulatory Science and Innovation (CERSI) grant to University of California, San Francisco (UCSF) and Stanford University, U01FD005978 funded by FDA/HHS. The contents are those of the author(s) and do not necessarily represent the official views of, nor an endorsement, by FDA/HHS, or the U.S. Government.

ORCID iDs: Betty Xiong (0000-0002-2047-7696), Jillian Fisher (0009-0005-9132-9374), Benjamin Newman (0000-0003-3552-8676), Meng Hu (0000-0003-2719-3263), Shivangi Gupta (0009-0007-7514-012X), Yejin Choi (0000-0003-3032-5378), Lanyan Fang (0000-0001-5378-2948), Russ Altman (0000-0003-3859-2905).

\bibliography{custom}

\clearpage
\appendix
\section{Additional results and analyses}
\label{app:additional_results}

\subsection{Evaluation and human annotation details}
\label{app:eval_details}

We use a combination of FDA experts and computer science annotators to validate our benchmark creation: FDA regulatory reviewers assessed \textbf{question} relevance and regulatory usefulness, while CS/biomedical data science annotators evaluated QA correctness and question quality using structured label-grounding rubrics.

The FDA expert evaluates complex in vivo and in vitro bioequivalence data for generic drug applications. They must hold professional degree in a relevant scientific field--such as Pharmacokinetics or Pharmaceutical Sciences--and possess extensive experience in designing and analyzing human pharmacokinetic studies. Furthermore, these assessors act as recognized subject matter experts who utilize modeling and simulation, develop regulatory standards, and mentor junior staff on intricate scientific issues. Drug labeling is one critical information source for their daily work.

We had three such domain judges to annotate a question's relevance to FDA drug review (Table~\ref{tab:expert_agreement_gpt_validator}). First, we ask the reviewers whether a question is relevant or irrelevant, and match it to the LLM's assessment. Second, we ask the assessors to categorize the question into one predefined regulatory topic category reflecting standard FDA label organization. Question topics can be found in Appendix~\ref{appendix:filtering}.

We had three computer science and biomedical data science annotators evaluate QA correctness and question quality, and compared their annotations to LLM-as-judge outputs. Table~\ref{tab:agreement_summary_cons_vs_unanim_cols} summarizes the agreement between human and LLM judgments in human consensus labels and the unanimous subset. Table~ \ref{tab:agreement_summary_ranges} shows annotator variability and human inter-annotator reliability.
Similarly, we had the above annotators evaluate answer correctness, and compared it against our SimpleQA prompt (modified for the specific FDA drug label task). Human and LLM agreement, and human annotator variability are reported in Tables~\ref{tab:agreement_summary_cons_vs_unanim_cols} and \ref{tab:agreement_summary_ranges}.

All LLM prompt details can be found in Appendix~\ref{app:prompts}.

\begin{table}[t]
\centering
\small
\begin{tabular}{lr}
\toprule
 \textbf{Metric} & \textbf{Value} \\
\midrule
\multicolumn{2}{c}{Binary human expert inter-rater agreement}\\
\midrule
 Mean pairwise agreement & 0.880\\
 Fleiss' $\kappa$ & 0.719\\
 Mean pairwise Cohen's $\kappa$ & 0.719\\
 Krippendorff's $\alpha$&0.720\\
\midrule
\multicolumn{2}{c}{Binary GPT validator vs. expert-majority reference}\\
\midrule
 Precision / PPV & 0.756\\
 Recall & 0.971\\
 F1 & 0.850\\
 Accuracy & 0.760\\
\bottomrule
\end{tabular}
\caption{
Expert agreement and GPT-validator audit on a manually reviewed subset of 50 FDA drug-label questions.
Binary relevance collapses ``Yes'' and ``Maybe'' into acceptable/retainable questions and treats ``No'' as reject.
Expert-majority decisions were used as the reference standard for evaluating the GPT validator; ties were resolved by random vote.
}
\label{tab:expert_agreement_gpt_validator}
\end{table}

\begin{table*}
\centering
\small
\setlength{\tabcolsep}{4pt}
\begin{tabular}{l l c c c c c c c c c c}
\hline
Stage / Judge& Task& \multicolumn{5}{c}{Consensus}
& \multicolumn{5}{c}{Unanimous} \\
& & $n$ & Accuracy& F1& Prec. & Rec.
  & $n$ & Accuracy& F1& Prec. & Rec. \\
\hline
QA correctness (LLM judge) & Factual
  & 19 & 0.684 & 0.750& 0.900 & 0.643
  & 10 & 0.900 & 0.933& 1.000 & 0.875 \\
QA correctness (LLM judge) & Multi-hop
  & 20 & 0.550 & 0.690& 1.000 & 0.526
  &  9 & 0.778 & 0.875& 1.000 & 0.778 \\
QA correctness (LLM judge) & Refusal
  & 10 & 0.700 & 0.769& 1.000 & 0.625
  &  6 & 0.833 & 0.909& 1.000 & 0.833 \\
\hline
Question quality (LLM judge) & Factual
  & 20 & 0.400 & 0.455& 0.833 & 0.312
  & 14 & 0.357 & 0.526& 0.833 & 0.385 \\
Question quality (LLM judge) & Multi-hop
  & 20 & 0.700 & 0.400& 0.333 & 0.500
  & 12 & 0.750 & 0.400& 0.333 & 0.500 \\
\hline
Answer correctness (SimpleQA) & Factual
  & 19 & 0.789 & 0.800& 1.000 & 0.667
  & 15 & 0.933 & 0.941& 1.000 & 0.889 \\
Answer correctness (SimpleQA) & Multi-hop
  & 15 & 0.733 & 0.667& 0.800 & 0.571
  &  8 & 0.750 & 0.750& 0.750 & 0.750 \\
\hline
\end{tabular}
\caption{LLM and judge agreement. Automated judges compared against (i) human consensus labels and (ii) the unanimous subset (all human annotators agree). Accuracy is percent agreement. F1, precision and recall treat the judge as predictor and the human labels as gold.}
\label{tab:agreement_summary_cons_vs_unanim_cols}
\end{table*}

\begin{table*}[h]
\centering
\tiny
\begin{tabular}{l l c c c c cll}
\hline
Stage / Judge & Task & $n$ & \multicolumn{4}{c}{Annotator variability}& \multicolumn{2}{c}{Human inter-annotator reliability}\\
& & & Accuracy range& F1 range& Prec. range & Rec. range  & Pairwise match range &Pairwise F1 range\\
\hline
QA correctness (LLM judge) & Factual  & 19 & 0.684--0.737 & 0.700--0.783 & 0.700--1.000 & 0.625--0.700  & 0.579--0.737 &0.692--0.828 \\
QA correctness (LLM judge) & Multi-hop & 20 & 0.500--0.750 & 0.643--0.762 & 0.800--1.000 & 0.500--0.727  & 0.500--0.850 &0.667--0.919 \\
QA correctness (LLM judge) & Refusal  & 10 & 0.600--0.800 & 0.714--0.833 & 1.000--1.000 & 0.556--0.714  & 0.600--0.900 &0.750--0.941 \\
\hline
Question quality (LLM judge) & Factual  & 20 & 0.400--0.400 & 0.455--0.455 & 0.833--0.833 & 0.312--0.312  & 0.700--0.900 &0.812--0.938 \\
Question quality (LLM judge) & Multi-hop & 20 & 0.650--0.700 & 0.250--0.533 & 0.167--0.667 & 0.400--0.500  & 0.650--0.850 &0.364--0.571 \\
\hline
Answer correctness (SimpleQA) & Factual  & 19 & 0.789--0.842 & 0.800--0.842 & 1.000--1.000 & 0.667--0.727  & 0.800--0.850 &0.833--0.880 \\
Answer correctness (SimpleQA) & Multi-hop & 15 & 0.467--0.800 & 0.429--0.727 & 0.600--0.800 & 0.333--0.667  & 0.500--0.875 &0.500--0.857 \\
\hline
\end{tabular}
\caption{Annotator variability and human inter-annotator reliability. First, the range across individual human annotators when compared against the automated judge (min--max over annotators). Second, pairwise ranges over all annotator pairs for each task.}
\label{tab:agreement_summary_ranges}
\end{table*}

\subsection{Error analysis on full-label setting}
\label{app:error_analysis}

To investigate the performance gap between full-label and oracle-passage, we perform a structural analysis on factual questions, averaging full-label correctness across models for each label and correlating label-level performance with document structure. We find that factual correctness decreases substantially as labels become longer and structurally more complex: label length (token) correlates negatively with performance (Spearman r = -0.57, p = 7.78e-41), as does number of passages (r = -0.45, p = 2.64e-26) and number of detected tables (r = -0.35, p = 5.72e-14). %

To investigate refusal errors and distinguish between cases where information is absent and cases where models decline due to complex reasoning requirements, we report false-refusal rates separately for factual and multi-hop questions (Table~\ref{tab:refusal_error}).

\begin{table*}[h]
    \centering
    \small
    \setlength{\tabcolsep}{4pt}
    \begin{tabular}{lrrrr}
\toprule
         Model&  Factual false refusal rate&  MH false refusal rate&  Correct abstention rate& Hallucination rate
\\
\midrule
         Llama-8B&  0.133&  0.179&  0.978& 0.022
\\
         Llama-70B&  0.150&  0.193&  0.982& 0.018
\\
 Ministral-14B& 0.196& 0.373& 0.994&0.006
\\
         Qwen3-14B&  0.208&  0.348&  0.997& 0.003
\\
         Qwen3-32B&  0.169&  0.208&  0.961& 0.039
\\
         Claude Opus&  0.111&  0.174&  0.774& 0.226
\\
         Claude Sonnet&  0.169&  0.341&  0.993& 0.007
\\
         GPT-4o-mini&  0.203&  0.331&  0.997& 0.003
\\
         GPT-5.1&  0.197&  0.308&  0.993& 0.007
\\
         GPT-5.2&  0.207&  0.366&  0.987& 0.013
\\
AVERAGE& 0.174& 0.282& 0.966&0.034\\
 \bottomrule
    \end{tabular}
    \caption{Error analysis for refusal.}
    \label{tab:refusal_error}
\end{table*}

\subsection{Comparison to other reasoning benchmarks}
\label{app:benchmark_comparison}

We computed Pearson correlations between multi-hop scores and scores on established benchmarks across the models used in our study. We also computed a linear fit to measure how well performance on our benchmark predicted performance on known benchmarks (Table~\ref{tab:comparison}). This analysis was conducted across ten models, and correlations should be interpreted with appropriate caution given the limited sample size (at N=10, only correlations exceeding approximately R = 0.63 reach conventional significance thresholds). With that caveat, the pattern of correlations is informative. MH shows positive correlations with HLE (R = 0.12) and CritPt (R = 0.21), both of which reward novel reasoning under conditions that resist memorization and retrieval. The negative correlation with MedQA (R = -0.41) is particularly telling: despite both benchmarks operating in the medical domain, MedQA rewards memorized clinical knowledge whereas MH requires cross-passage inference from provided text, suggesting the two tap distinct capabilities. Near-zero correlations with agentic benchmarks such as GDPval-AA further confirm that MH performance is not driven by general task-completion or domain knowledge ability. %

\begin{table*}[h]
\small
    \centering
    \begin{tabular}{lllll}
    \toprule
         Benchmark Type&  Benchmark&  R&  Slope& Intercept
\\
\midrule
         Reasoning (scientific)&  HLE Score&  0.115&  37.4638& -0.0778
\\
         Reasoning (scientific)&  CritPt Score&  0.207&  25.8286& -6.9354
\\
         Domain Specific (medical)&  MedQA Score&  -0.4054&  -115.4889& 139.8816
\\
         Agentic (real world tasks)&  GDPval-AA Score&  -0.0152&  -8.4633& 27.2843\\
\bottomrule
    \end{tabular}
    \caption{Pearson correlation between multi-hop scores and other reasoning benchmarks.}
    \label{tab:comparison}
\end{table*}

\subsection{Details on selected models}

Tables \ref{tab:retrievers} and \ref{tab:models} show details on selected retriever and LLM models selected for evaluation.

\begin{table*}[h]
\centering
\small
\begin{tabular}{llll}
\toprule
\textbf{Retriever} & \textbf{Type} & \textbf{Representation / Scoring}  &\textbf{Reference}\\
\midrule
BM25 & Sparse & Token-based ranking over SPL passages  & \cite{bm25}\\
Cosine Similarity & Dense (embeddings) & Cosine similarity over passage embeddings  & \cite{all-MiniLM-L6-v2}\\
ReContriever & Dense & Learned dense retriever over passages  & \cite{recontriever}\\
\bottomrule
\end{tabular}
\caption{Retrievers evaluated for top-$k$ evidence selection in retrieval-augmented QA.}
\label{tab:retrievers}
\end{table*}

\begin{table*}[h]
\centering
\small
\begin{tabular}{llll}
\toprule
\textbf{Family} & \textbf{Model} & \textbf{Access}  &\textbf{Reference}\\
\midrule
Proprietary & GPT-4o-mini & API  & \cite{openai2024gpt4ocard}\\
Proprietary & GPT-5.1 & API  & \cite{singh2025openaigpt5card}\\
 Proprietary & GPT-5.2& API  &\cite{gpt5.2}\\
Proprietary & Claude-Sonnet-4-5 & API  & \cite{claude}\\
 Proprietary & Claude-Opus-4-6& API  &\cite{claude_opus}\\
\midrule
Open-weight & Llama-3.1-8B-Instruct & Local  & \cite{grattafiori2024llama3herdmodels}\\
Open-weight & Llama-3.3-70B-Instruct & Local  & \cite{grattafiori2024llama3herdmodels}\\
Open-weight & Ministral-3-14B-Instruct-2512 & Local  & \cite{ministral3}\\
Open-weight & Qwen3-14B & Local  & \cite{qwen3}\\
Open-weight & Qwen3-32B & Local  & \cite{qwen3}\\
\bottomrule
\end{tabular}
\caption{Models evaluated in our experiments.}
\label{tab:models}
\end{table*}

\subsection{Text metrics and additional plots}
\begin{figure}[h]
    \centering
    \includegraphics[width=\linewidth]{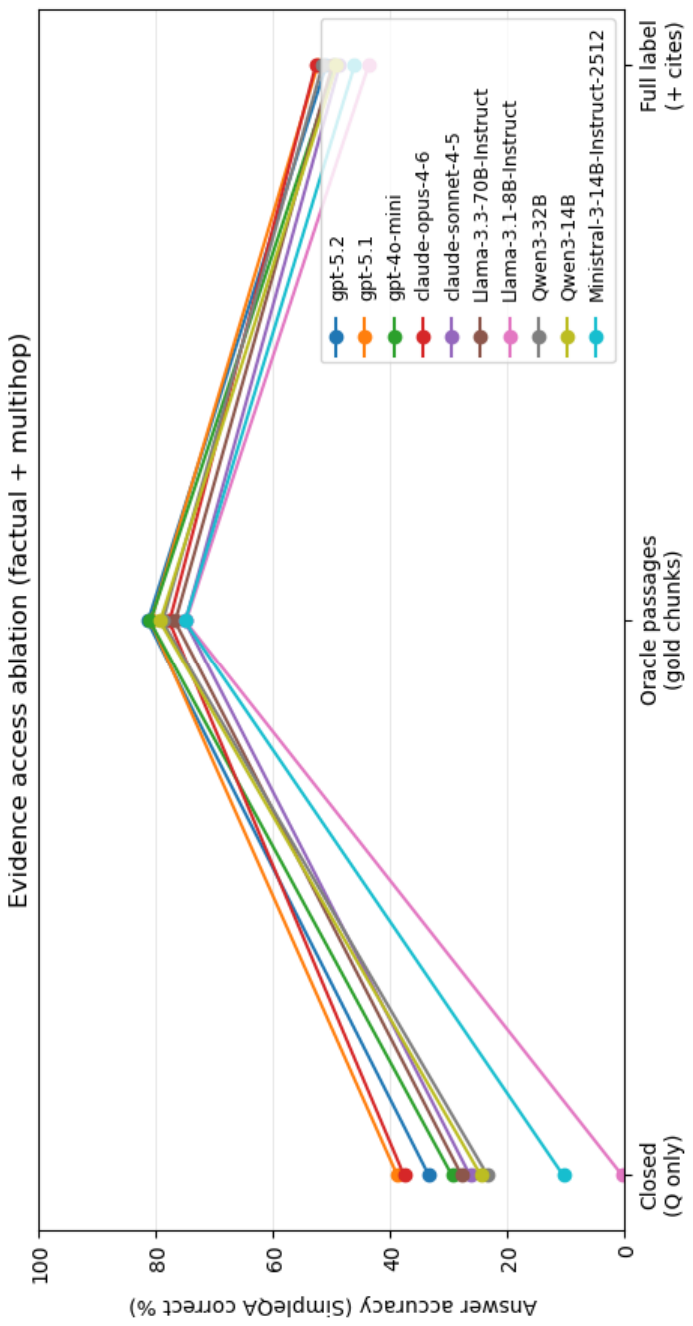}
    \caption{Evidence access ablation (factual + multi-hop). Answer accuracy improves substantially when models are given oracle (gold) passages, but drops in the full-label setting with citation requirements.} %
    \label{fig:ablation}
\end{figure}

\begin{figure}[h]
    \centering
    \includegraphics[width=\linewidth]{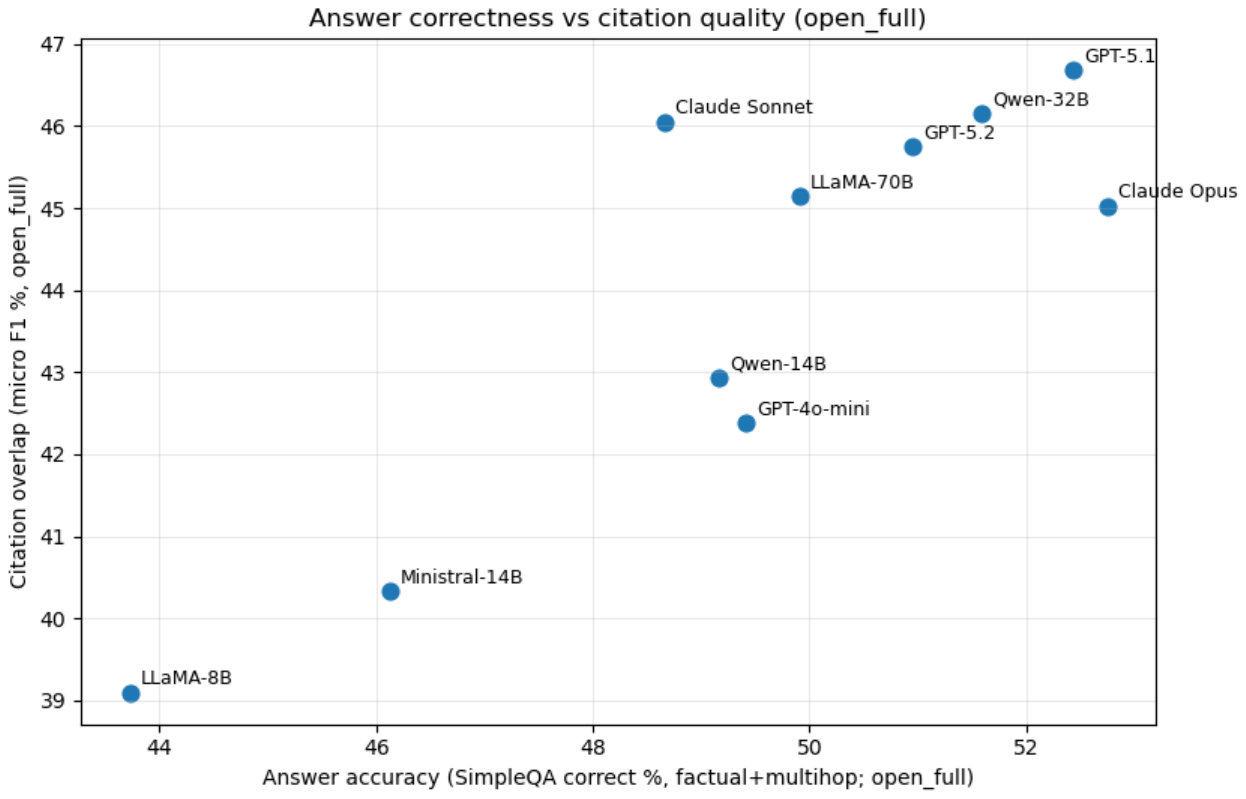}
    \caption{Answer correctness vs. citation quality in full-label setting. Relationship between overall answer accuracy and citation overlap (micro-F1).} %
    \label{fig:citation}
\end{figure}

\begin{figure}[h]
    \centering
    \includegraphics[width=\linewidth]{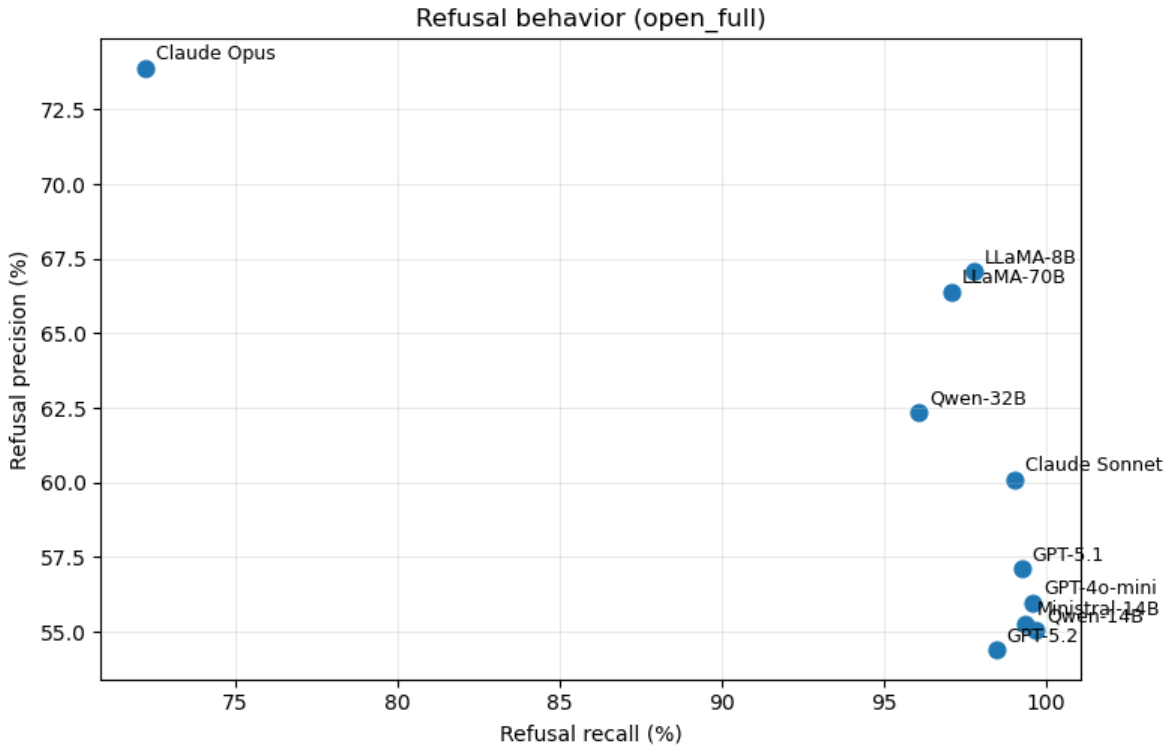}
    \caption{Refusal behavior in full-label setting. Models vary in hallucination resistance despite high recall.}
    \label{fig:refusal}
\end{figure}

Table \ref{tab:text_metrics_by_evidence_setting} shows additional analysis, with \textit{BLEU}, \textit{METEOR}, and \textit{ROUGE-L}. Figures~\ref{fig:ablation}, \ref{fig:citation} and \ref{fig:refusal} show additional analysis on model performance.

\begin{table*}[h]
\centering
\tiny
\setlength{\tabcolsep}{1.25pt}
\renewcommand{\arraystretch}{1.0}
\begin{tabular}{l|ccc|ccc|ccc|ccc|ccc|ccc}
\toprule
\textbf{Model} &
\multicolumn{3}{c|}{\textbf{Closed-book Fact}} &
\multicolumn{3}{c|}{\textbf{Closed-book MH}} &
\multicolumn{3}{c|}{\textbf{Oracle passages Fact}} &
\multicolumn{3}{c|}{\textbf{Oracle passages MH}} &
\multicolumn{3}{c|}{\textbf{Full label Fact}} &
\multicolumn{3}{c}{\textbf{Full label MH}} \\
&
\textbf{BLEU} & \textbf{MET} & \textbf{R-L} &
\textbf{BLEU} & \textbf{MET} & \textbf{R-L} &
\textbf{BLEU} & \textbf{MET} & \textbf{R-L} &
\textbf{BLEU} & \textbf{MET} & \textbf{R-L} &
\textbf{BLEU} & \textbf{MET} & \textbf{R-L} &
\textbf{BLEU} & \textbf{MET} & \textbf{R-L} \\
\midrule
Llama-8B  &
0.03 & 0.11 & 0.12 &
0.03 & 0.11 & 0.13 &
0.33 & 0.58 & 0.56 &
0.21 & 0.47 & 0.43 &
0.23 & 0.40 & 0.41 &
0.12 & 0.28 & 0.30 \\
Llama-70B &
0.07 & 0.27 & 0.25 &
0.10 & 0.36 & 0.30 &
0.34 & 0.59 & 0.57 &
0.23 & 0.50 & 0.45 &
0.23 & 0.43 & 0.41 &
0.16 & 0.37 & 0.34 \\
Ministral-14B &
0.03 & 0.20 & 0.18 &
0.04 & 0.28 & 0.23 &
0.16 & 0.43 & 0.50 &
0.10 & 0.38 & 0.37 &
0.18 & 0.37 & 0.39 &
0.10 & 0.27 & 0.26 \\
Qwen-14B &
0.08 & 0.29 & 0.27 &
0.12 & 0.37 & 0.33 &
0.33 & 0.60 & 0.56 &
0.24 & 0.54 & 0.47 &
0.21 & 0.41 & 0.39 &
0.14 & 0.31 & 0.28 \\
Qwen-32B &
0.08 & 0.29 & 0.26 &
0.10 & 0.38 & 0.31 &
0.31 & 0.60 & 0.55 &
0.21 & 0.54 & 0.45 &
0.22 & 0.42 & 0.41 &
0.16 & 0.37 & 0.34 \\
Claude Sonnet &
0.06 & 0.30 & 0.22 &
0.06 & 0.35 & 0.24 &
0.31 & 0.62 & 0.55 &
0.17 & 0.51 & 0.40 &
0.21 & 0.45 & 0.40 &
0.12 & 0.34 & 0.28 \\
GPT-4o-mini &
0.08 & 0.29 & 0.26 &
0.12 & 0.38 & 0.33 &
0.36 & 0.62 & 0.58 &
0.27 & 0.57 & 0.50 &
0.21 & 0.42 & 0.39 &
0.16 & 0.35 & 0.32 \\
GPT-5.1 &
0.09 & 0.30 & 0.28 &
0.07 & 0.33 & 0.28 &
0.30 & 0.58 & 0.56 &
0.17 & 0.48 & 0.41 &
0.21 & 0.41 & 0.41 &
0.12 & 0.32 & 0.29 \\
\bottomrule
\end{tabular}
\caption{Text-overlap metrics across evidence settings. Entries report \emph{BLEU}, \emph{METEOR (MET)}, and \emph{ROUGE-L (R-L)} (higher is better) for factual (Fact) and multi-hop (MH) questions in the closed-book, oracle-passages, and full-label settings.}
\label{tab:text_metrics_by_evidence_setting}
\end{table*}

 \clearpage
\clearpage
\section{Dataset creation details}
\label{appendix:dataset}

\subsection{Source Documents and Preprocessing}
\label{appendix:source-docs}

To build the dataset, we curate a set of drug label SPL XML documents from the FDALabel Database \cite{fdalabeldb}. Like similar works \cite{askfdalabel, unitox}, we start with the set of all human prescription drugs. We remove labels whose route of administration include the words topical, irrigational, intradermal or inhalation, as we are mostly interested in drugs with oral administration. We randomly sampled 700 drug labels for the benchmark.

Each SPL has a hierarchical structure that includes a mandatory \textit{Highlights of Prescribing Information} section (which is not included in the main drug label retrieval), PLR-mandated labeled sections, optional subsections, and structured metadata. We parse each XML file into section-level text chunks, using a custom rule-based extractor that preserves the hierarchical structure while normalizing content into contiguous natural-language segments. For each SPL, we extract sections with: (1) section titles, (2) LOINC section codes (e.g., 34084-4 for \textit{Adverse Reactions}), (3) chunk indices (at document and section levels) for provenance and citation tracking, and (4) drug-level metadata. Text normalization removes XML artifacts, formatting tags, superscripts, hyperlinks, and tables that cannot be reliably converted to plain text. Very short fragments created by XML segmentation (typically originating from bullet lists, warnings, or embedded bold headers) are merged with adjacent segments to produce semantically coherent units. The resulting corpus contains tens of thousands of section-level chunks that serve as the base context for question generation.

\subsection{Question Types}
\label{appendix:question_types}

Table~\ref{tab:example_items} shows examples of factual, multi-hop and refusal questions.

\begin{table*}[t]
\centering
\small
\setlength{\tabcolsep}{6pt}
\renewcommand{\arraystretch}{1.2}
\begin{tabular}{p{1cm} p{2.5cm} p{4cm} p{6cm}}
\toprule
\textbf{Type} & \textbf{Description} & \textbf{Example Question} & \textbf{Gold Target / Evidence (excerpt)} \\
\midrule
\textbf{Factual} &
Answerable from a single section &
What is the recommended starting dose of WAKIX for adult patients? &
\textbf{Gold:} The recommended starting dose is 8.9 mg once daily. \newline
\textbf{Evidence:} \emph{Adult Patients: Initiate WAKIX at 8.9 mg once daily and titrate to a maximum recommended dosage of 17.8 mg once daily after 7 days.} \\
\midrule
\textbf{Multi-hop} &
Requires reasoning across sections &
What is the peak blood level of dextroamphetamine achieved after ingesting 10 mg of the oral solution compared to the available strength of the oral solution? &
\textbf{Gold:} The peak blood level of dextroamphetamine after ingesting 10 mg of the oral solution is 33.2 ng/mL, while the available strength of the oral solution is 5 mg/5 mL. \newline
\textbf{Evidence:} \emph{...produced an average peak dextroamphetamine blood level of 33.2 ng/mL...} \newline
\emph{Dextroamphetamine Sulfate Oral Solution 5 mg/5 mL...} \\
\midrule
\textbf{Refusal} &
Not answerable from the label; model should abstain &
How should BNP be monitored in patients taking New Day? &
\textbf{Gold:} NOT\_ANSWERABLE \\
\bottomrule
\end{tabular}
\caption{Example items from the benchmark spanning factual, multi-hop, and refusal question types. Evidence excerpts are abbreviated for space.}
\label{tab:example_items}
\end{table*}

\paragraph{Factual.}  
The FDA Highlights section provides a concise summary of the most essential clinical information in a drug label, including indications, dosage forms, contraindications, boxed warnings, and key safety considerations. Since the highlights are carefully curated and tightly coupled to the corresponding full sections of the SPL, it serves as a high-quality but weakly-supervised source of factual supervision. For each highlight in the label, we derive the corresponding full-length sections. We then use the section header to select from a predefined library of template-based questions tailored to common regulatory information needs. For example, from the \textit{Indications and Usage} heading, we generate question templates such as ``What is \{drug\} used to treat?'' or ``What conditions is \{drug\} indicated for?''; from \textit{Dosage and Administration}, we generate templates such as ``What is the recommended dosage of \{drug\}?''. As each question is explicitly anchored to both the highlights summary and its corresponding extractable full section span, these items provide a reliable factual baseline for evaluation.

To capture a broader range of clinically and regulatory meaningful content beyond the Highlights summary, we generate section-level factual questions using a self-instruct prompting strategy. For each drug label, we iterate through individual sections and select one chunk at a time as the input context. Each chunk is annotated with its section identifier and document chunk index to preserve provenance. Using few-shot prompts to an LLM (GPT-4o-mini \footnote{Unless otherwise specified, all generation and filtering prompts use GPT-4o-mini.} \cite{openai2024gpt4ocard}), we instruct the model to propose clinically relevant, factually grounded questions whose answers can be derived directly from the provided text. Multiple QA items are generated per chunk to enhance coverage. This method captures a richer set of real-world information needs, including dose adjustments, contraindicated populations, drug-drug interactions, adverse event frequencies, and pharmacokinetic properties. Since the questions originate from individual sections rather than curated summaries, they reflect the full variability of label writing styles and provide a challenging and realistic substrate for factual QA evaluation.

\paragraph{Multi-hop.}  
Many regulatory and clinical information needs require synthesizing evidence across multiple sections of a drug label. To evaluate this capability, we construct multi-hop QA items requiring integration of two distinct sections. We begin by sampling pairs of sections that often interact clinically, e.g., \textit{Indications and Usage} paired with \textit{Dosage Forms and Strengths}, or \textit{Pharmacokinetics} paired with \textit{Adverse Reactions}. For each selected pair from a single drug label, we retrieve one chunk from each section, concatenate their texts, and provide both as context to an LLM. The model is explicitly instructed to generate questions whose answers require simultaneous reasoning over both sections and cannot be answered from either section alone. Valid multi-hop questions that genuinely depend on both sections are retained and subsequently evaluated using a judge model that enforces cross-sectional evidence requirements. This category serves as a stress test for cross-contextual reasoning and evidence integration in long, heterogeneous biomedical documents.

Despite these constraints, initial generations often produce questions that reference unrelated medical concepts or introduce unsupported associations (e.g., ``What is the breast cancer risk associated with long-term use of CAMRESE, and how is it supplied?'' or ``Why must PROGESTERONE Capsules be avoided in patients with a peanut allergy despite its renal excretion of metabolites?''). Such items are removed during downstream filtering. 

\paragraph{Refusal.}  
Safe model behavior requires not only producing correct answers when information is present but also refusing to answer when information is absent. To evaluate hallucination resistance, we construct a set of negative-control refusal questions referencing biomedical concepts absent from the label. We begin with a library of template-based prompts commonly used in clinical and regulatory settings, such as ``What is the indication of \{drug\} for treating \{endpoint\} in \{population\}?'' or ``What is the threshold value of \{biomarker\} for initiating treatment with \{drug\}?''. Into these templates, we insert a biomedical keyword, entity, or condition that is guaranteed to be absent from the SPL. We verify absence automatically through full-label keyword search. To improve naturalness and surface variability, each template is then rephrased by an LLM, producing more linguistically diverse unanswerable questions while preserving the inserted out-of-scope entity. The correct system response must decline to answer or explicitly state that the label does not contain relevant information. These items provide a clean measure of hallucination propensity under controlled distribution shift. This category is particularly important for regulatory applications, where unsupported claims pose significant safety and compliance risks.

\subsection{Generation Pipeline}
\label{appendix:generation}

\paragraph{Context Selection.}
We choose the appropriate text source depending on the question type. Highlights produce template-driven factoids; single sections support self-instruct factual items; and section pairs enable multi-hop reasoning. All metadata, such as drug name, section identifiers, and chunk indices, is preserved to ensure traceability. This is particularly important for the task of document-level citations that is part of the evaluation suite in an open-book full drug label setting (see Appendix \ref{appendix:full_label}).

\paragraph{Prompted Generation.}
We craft task-specific few-shot prompts that emphasize regulatory precision, prohibition of hallucination, and clarity of reasoning. For self-instruct factual and multi-hop items, we ask the LLM to create question-answer pairs specifically relevant to only the specified context(s). For multi-hop questions, prompts stress that the answer must require information from both sections.

For refusal questions, prompts reinforce that the model should avoid inventing content and instead explicitly refuse. Each generation round produces multiple independent samples to enhance diversity. 

\paragraph{Expert-Guided Stage.} 
Domain experts (FDA drug label reviewers) annotate a seed set of QAs and provide generalizable constraints, e.g., determining regulatory and clinical relevance of questions, importance of traceability and provenance, compound multi-hop questions relevant that may include maximum daily dose in certain patient populations. We encode this feedback into refined prompts and filtering criteria for improving benchmark examples.

\subsection{Filtering and Quality Control}
\label{appendix:filtering}

\paragraph{Stage 1: Rule-Based Filtering.}  
We first apply a set of deterministic, rule-based filters designed to eliminate structurally invalid, irrelevant, or weakly grounded QAs before any model-based judgment. These rules encode minimal regulatory and linguistic constraints that can be verified without semantic interpretation, ensuring transparency and reproducibility. 

Structural validity checks discard QAs missing any required field (question, context, or answer) or exhibiting malformed serialization. Each question is required to explicitly mention the target drug by name or known alias, preventing generic or context-free questions from entering the benchmark. We enforce length guards on both questions (10-200 characters) and answers (5-600 characters), rejecting items that fall outside reasonable minimum and maximum thresholds in order to remove trivial, underspecified prompts as well as excessively verbose or malformed generations.

For multi-hop candidates, we enforce additional structural constraints to prevent degenerate single-hop formulations. We reject questions containing artificial conjunctions such as ``and'' or ``as well as'' that merely concatenate two unrelated facts without requiring integrative reasoning, collapsing into separate single-hop formulations, e.g., ``What is the breast cancer risk associated with long-term use of CAMRESE, and how is it supplied?''. This filter removes a common failure mode in which the model produces compound questions that are answerable independently from each section. 

Finally, we apply span-based support checks to ensure extractive grounding. For factual questions, the answer must exhibit a minimum token-level overlap of 0.10 with the provided context. For multi-hop questions, we require a minimum overlap of 0.10 with each constituent section independently. Items failing any of these criteria are deterministically discarded prior to LLM-based validation.

\paragraph{Stage 2: LLM-as-Judge Validation.}  
After rule-based filtering, we apply a structured LLM-as-judge pipeline to perform semantic validation of remaining QAs. This stage is responsible for assessing factual correctness, evidentiary grounding, and regulatory relevance using only the provided label context. To ensure interpretability and auditability, the LLM-based filtering is decomposed into three explicit judgment stages, each producing structured outputs that are logged and retained for downstream analysis.

\paragraph{(i) Question Relevance Classification.}  
In the second stage, the judge evaluates whether the question itself is relevant to FDA regulatory review, independent of answer correctness. Each question is classified into exactly one predefined regulatory topic category reflecting standard FDA label organization:
\begin{enumerate}
    \item Indications \& Usage
    \item Dosage \& Administration
    \item Dosage Forms \& Strengths \& Formulation
    \item Contraindications
    \item Safety \& Serious Risks
    \item Adverse Events
    \item Drug Interactions
    \item Use in Specific Populations
    \item Pharmacokinetics
\end{enumerate}
Questions that do not correspond to regulatory labeling content are assigned to a \textit{None / Irrelevant} category. In addition, the judge labels each question as relevant, somewhat relevant, or irrelevant to bioequivalence review. Questions deemed irrelevant or only weakly related to regulatory decision-making are filtered out at this stage, ensuring topical alignment of the final benchmark.

\paragraph{(ii) QA Validation and Evidence Extraction.}  
In the first stage, the judge evaluates whether the proposed answer is supported by the provided context text alone, ignoring any external knowledge. For factual questions, the judge classifies each item as supported or unsupported and, when supported, extracts the minimal sentence-level evidence from the label that directly substantiates the answer. For multi-hop questions, the judge independently evaluates support from each section, producing separate fields for evidence from Context~A and Context~B, and explicitly determines whether both sections are required to form a complete answer. Questions whose answers are supported by only one section, despite being framed as multi-hop, are penalized and removed. For refusal questions, the judge verifies whether the absence-of-information claim is correct; if the label does in fact contain an answer, the judge extracts the contradicting evidence span. This stage ensures that retained QAs are factually grounded and traceable to specific passages in the source document.

\paragraph{(iii) Question Quality Assessment.}  
In the final stage, the judge evaluates the intrinsic quality of each question using task-specific criteria. For factual questions, the judge flags failures such as answer leakage (where the question states its own answer), context mismatch (mentioning unsupported entities, populations, or conditions), unanswerability from the provided context, or low benchmarking utility due to vagueness or triviality. For multi-hop questions, additional failure modes are assessed, including one-sided answerability, artificial or nonsensical hop construction between sections, answer leakage from one context into the question, and inability to answer using the combined contexts. Questions exhibiting any of these failure modes are removed. This stage ensures that retained questions are not only factually valid but also meaningful and well-formed evaluation items.

Together, these three LLM-based filtering stages provide a conservative but interpretable semantic validation layer that complements the deterministic rule-based filters. By separating correctness, relevance, and question quality judgments, the pipeline enables fine-grained auditing and supports scalable, reproducible benchmark construction under regulatory constraints.

\paragraph{Precision-Oriented Filtering Rationale.}  
We explicitly tune the LLM-based filtering stage to maximize precision rather than recall. Human annotation studies conducted on stratified samples of factual, multi-hop, and refusal items show that while human--human agreement is moderate, reflecting the inherent ambiguity of regulatory QA, the LLM judge achieves consistently high precision with respect to human consensus, albeit with lower recall. In practice, this means the filter preferentially excludes borderline or ambiguous items that some humans might accept, while rarely admitting incorrect examples. We consider this trade-off desirable for benchmark construction, as omission of marginal cases is preferable to inclusion of incorrect or weakly grounded QAs in high-stakes regulatory evaluation.

\paragraph{Sampling for Human Review and Auditability.}  
To support human auditing and agreement analysis, we construct a reproducible sampling pipeline over filtered QAs. Items are deduplicated by unique question identifiers, bucketed by regulatory category and relevance label, and sampled to ensure balanced coverage across question types and sections. For each sampled item, we export structured review files containing the question, context, model answer, judge decision, and metadata required for downstream human evaluation. This infrastructure enables systematic comparison between human judgments and LLM-based filtering decisions and provides an auditable trail for benchmark curation.

\paragraph{Final Dataset.}  
After completion of all filtering stages, the benchmark contains \textit{17223} high-quality QAs across three categories (see Table \ref{tab:dataset_stats}). Detailed statistics are reported in Appendix~\ref{appendix:dataset-stats}. The resulting dataset reflects a conservative but reliable subset of regulatory QA examples, suitable for evaluating factual grounding, cross-section reasoning, and safe refusal behavior in large language models.

\subsection{Dataset Statistics and Filtering Diagnostics}\label{appendix:dataset-stats}

\begin{figure}[h]
    \centering
    \includegraphics[width=\linewidth]{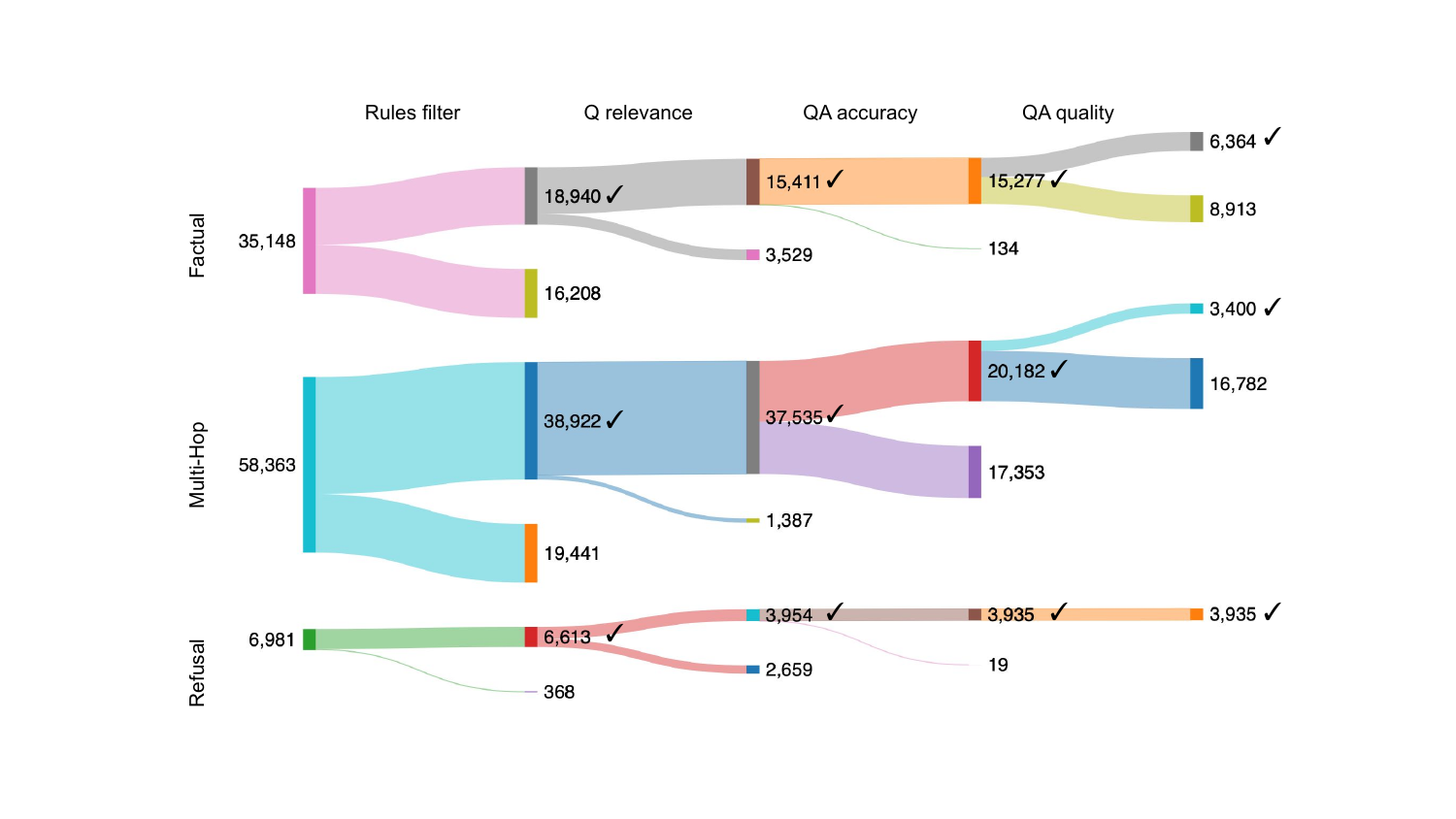}
    \caption{Sankey plot of retention of document QA items across filtering stages. \ding{51} denotes number of items kept after each stage.}
    \label{fig:sankey}
\end{figure}

See Table~\ref{tab:dataset_stats} for dataset statistics. Figure~\ref{fig:sankey} shows a Sankey plot on the filtering steps in order of: rules-based automatic filter; question relevance, QA accuracy, and question quality LLM-based filters.

\begin{table}[t]
\centering
\small
\setlength{\tabcolsep}{10pt}
\renewcommand{\arraystretch}{1.15}
\begin{tabular}{l r}
\toprule
\textbf{Dataset statistic} & \textbf{Count} \\
\midrule
SPL drug labels & 700 \\
Final QAs & 17{,}223 \\
Factual & 9{,}888 \\
Multi-hop & 3{,}400 \\
Refusal & 3{,}935 \\
\bottomrule
\end{tabular}
\caption{Dataset summary statistics.}
\label{tab:dataset_stats}
\end{table}

\subsection{Benchmark Tasks}
\label{appendix:benchmark_tasks}
\paragraph{Closed-Book Question Answering.}
In the closed-book setting, models are provided with the question only and must generate an answer without access to any label context. This task evaluates parametric knowledge retention, robustness to hallucination, and a model’s tendency to overgeneralize from pretraining rather than abstain when uncertain. Closed-book QA is particularly informative for assessing whether models incorrectly rely on memorized drug knowledge that may be outdated, incomplete, or inapplicable to a specific formulation.

\paragraph{Open-Book Question Answering (Full Label with Citations).}
\label{appendix:full_label}
In the open-book full-label setting, models are provided with the entire SPL document divided into passages with explicit passage identifiers, e.g., \texttt{PASSAGE\_0005}. Models must generate (1) a concise natural-language answer and (2) a list of cited passage ids that support the answer. This setting evaluates long-context comprehension, evidence grounding, and provenance quality under realistic document length and redundancy. Requiring explicit citations enables automatic verification of whether the model’s stated evidence actually supports the answer and helps separate failures of retrieval/reading from failures of reasoning.

\paragraph{Retrieval Task for Open-Book Question Answering}
\label{appendix:retrieval_task}

To better reflect practical reviewer workflows and to decouple long-context reading from evidence selection, we also evaluate retrieval-augmented open-book QA. Given a question and a corpus of passage-level label chunks, a retriever selects the top-$k$ passages (we use $k{=}5$ unless otherwise stated). We experiment with multiple retrieval functions, including (1) sparse retrieval (BM25) \cite{bm25}, (2) embedding-based cosine similarity over passage representations, and (3) a dense retriever (ReContriever) \cite{recontriever}. This setting evaluates the task of  identifying relevant evidence prior to answer generation.

\paragraph{Section-Aware Question Answering}

To isolate reasoning from evidence selection, we include a section-aware setting where the model receives the question along with the specific section chunk(s) (``gold'' retrieval variant) from which the QA was generated. For self-instruct factual questions, the model receives a single section chunk; for multi-hop questions, the model receives exactly two section chunks corresponding to Context~A and Context~B. This diagnostic setting serves as an approximate upper bound on answer quality when the relevant evidence is perfectly localized in an oracle reference, and helps quantify how much error in open-book settings is attributable to retrieval/selection rather than reasoning.

\paragraph{Refusal Detection}

Finally, we evaluate refusal behavior as a dedicated negative control task, to assess a model’s ability to safely abstain from answering when the label does not contain the requested information. The expected behavior is to explicitly abstain (e.g., \texttt{NOT\_ANSWERABLE} or an equivalent refusal) rather than fabricate an answer. We track hallucinated answers as critical failures, and we also track false refusals (refusing despite the label containing relevant information) as an error mode that harms usability. This task is particularly important in regulatory settings, where unsupported claims can pose safety and compliance risks.

\subsection{Automatic Metrics}
\label{appendix:metrics}

\paragraph{Answer Quality.}
For answerable questions (factual, and multi-hop), we compute BLEU, METEOR, and ROUGE-L between the model prediction and the reference answer when a reference is available. These metrics provide a coarse measure of lexical similarity, but are known to be insufficient for capturing factual correctness in long-form biomedical QA due to multiple valid phrasings and levels of granularity. Accordingly, they are always reported in conjunction with LLM-as-judge scores (see Appendix \ref{appendix:judge}).

For questions where the gold answer is present but the model outputs a refusal (e.g., \texttt{NOT\_ANSWERABLE}), overlap-based metrics naturally assign low scores, and the LLM-as-judge marks the prediction as incorrect. This behavior appropriately penalizes over-refusal in answerable cases without requiring special handling.

\paragraph{Refusal Correctness.}
For refusal questions, evaluation is framed as a binary classification task: the model is correct if it explicitly refuses to answer when the information is absent from the label, and incorrect otherwise. We report precision, recall, and accuracy. Because hallucinated answers in refusal cases constitute a critical safety failure, we emphasize precision as the primary metric, while still reporting recall to quantify over-conservatism.

\paragraph{Provenance and Citation Evaluation.}
For open-book full-label and retrieval-based settings, models are required to produce explicit citations to supporting passages \cite{openscholar}. For each example, we compare the set of cited passages against the gold provenance associated with the QA generation process. We compute per-example precision, recall, and F1 over cited passage identifiers and report averages across the dataset.

We explicitly distinguish between false-positive citations (citing irrelevant passages) and false-negative citations (failing to cite required evidence). In regulatory contexts, missing critical evidence is often more harmful than citing additional, partially relevant passages. Accordingly, we report recall-oriented citation metrics alongside F1 to reflect the importance of evidence completeness during review.

\subsection{LLM-Based Grading Framework}
\label{appendix:judge}

Many long-form benchmarks items do not admit a single canonical reference answer, so we employ an LLM-as-judge framework. To evaluate model predictions at scale, we use an LLM-based grading framework (with GPT-5.1) that compares predicted answers against gold reference targets derived from FDA drug labels.

For each evaluation example, the grader is provided with (1) the question, (2) a gold target answer extracted from the label, and (3) the model’s predicted answer. The grader assigns one of three mutually exclusive labels:
\texttt{CORRECT}, \texttt{INCORRECT}, or \texttt{NOT\_ATTEMPTED}. A prediction is marked as correct if it contains all clinically important information present in the gold target, does not contradict the reference, and does not introduce unsupported clinical claims. Minor wording differences and paraphrases are permitted as long as the clinical meaning is preserved. Predictions are marked incorrect if they contradict the gold target, introduce unsupported clinical facts (e.g., incorrect doses, populations, contraindications), or omit major required elements. Predictions are marked as not attempted if the model explicitly abstains or fails to provide the requested information without introducing incorrect claims.

Building upon the SimpleQA evaluation framework \cite{simpleqa}, the grading rubric is designed to reflect regulatory priorities. In particular, missing or incorrect numeric values (e.g., doses, frequencies, thresholds), incorrect population constraints, and unsupported contraindications are treated as critical errors and result in an incorrect label. When the gold target specifies refusal or absence of information, any specific clinical recommendation from the model is graded as incorrect, while explicit abstention is graded as not attempted. This structured grading scheme enables consistent comparison across models while preserving clinically meaningful distinctions between incorrect answers and safe non-attempts.

We emphasize that the LLM judge is used as a scalable proxy for semantic validation rather than as a source of ground truth. Its outputs are therefore evaluated against human judgments in a separate annotation study in Appendix \ref{appendix:human_eval}.

\subsection{Human Annotation Study}
\label{appendix:human_eval}

To validate the reliability of automatic grading and to characterize ambiguity in regulatory QA, we conduct a human annotation study using the same grading rubric applied by the LLM-based evaluator. Annotators label each prediction as \texttt{CORRECT}, \texttt{INCORRECT}, or \texttt{NOT\_ATTEMPTED}, following identical criteria regarding clinical completeness, contradictions, and appropriate abstention.

Each example is independently annotated by three reviewers with biomedical or regulatory expertise. We measure inter-annotator agreement using Cohen's $\kappa$. We additionally report bootstrap confidence intervals over agreement statistics to quantify uncertainty due to sample size.

Because regulatory QA often involves borderline cases and underspecified label language, we do not expect near-perfect agreement. Instead, these measurements provide a realistic estimate of task ambiguity and establish an upper bound on achievable automatic grading consistency.

Overall, this evaluation framework combines automatic metrics, structured model-based judgment, and human validation to provide a robust and interpretable assessment of regulatory QA performance under both answerable and unanswerable conditions.

\clearpage

\subsection{Dataset creation prompts}
\label{app:prompts}

\paragraph{Generation prompts.}
Template for highlight-generated QA examples.

\begin{promptbox}
{
  "INDICATIONS AND USAGE": [
    "What is {drug} used to treat?",
    "What conditions is {drug} indicated for?"
  ],
  "DOSAGE AND ADMINISTRATION": [
    "What is the recommended dosage regimen of {drug}?",
    "How should {drug} be administered?"
  ],
  "DOSAGE FORMS AND STRENGTHS": [
    "What dosage forms are available for {drug}?",
    "What strengths does {drug} come in?"
  ],
  "CONTRAINDICATIONS": [
    "Who should not take {drug}?",
    "What are the contraindications for {drug}?"
  ],
  "WARNINGS AND PRECAUTIONS": [
    "What important warnings or precautions are associated with {drug}?",
    "What safety risks are listed for {drug}?"
  ],
  "ADVERSE REACTIONS": [
    "What are the side effects of {drug}?",
    "What adverse reactions have been reported for {drug}?"
  ],
  "DRUG INTERACTIONS": [
    "What important drug interactions are noted for {drug}?",
    "Which medications should be avoided with {drug}?"
  ],
  "USE IN SPECIFIC POPULATIONS": [
    "What is known about the use of {drug} in specific populations?",
    "Are there any population-specific considerations for {drug}?"
  ],
  "WARNING: ": [
    "What serious risks are included in the boxed warning for {drug}?",
    "What does the boxed warning for {drug} emphasize?"
  ]
}
\end{promptbox}
    
Prompt template for self-instruct generation of single-hop QA examples.

\begin{promptbox}
Generate {k} grounded Q/A items from the CONTEXT below. Requirements:
- Use ONLY facts in CONTEXT.
- Keep answers concise (< 2 sentences or a compact list).
- Questions and answers should be concrete and clearly useful for bioequivalence or product-specific guidances development.
- Output a JSON array ONLY with objects: {{"question":"...", "answer":"...", "type":"factual|list|numeric"}}
- Do NOT include citations or any extra fields.

CONTEXT (section: "{title}" / label: {label} / chunk_index: {chunk_index}):
{context}
\end{promptbox}

Prompt template for self-instruct generation of multi-hop QA examples.

\begin{promptbox}
Create ONE multi-hop item that needs BOTH contexts.

Rules:
- Single-clause; avoid " and ".
- One decision, not two unrelated facts.
- Each section must contribute a distinct needed fact.

Positive examples (style):
- (Dosage + Hepatic impairment) - "What starting dose is recommended for patients with mild hepatic impairment treated for angina?"
- (Indication + Contraindication) - "Although indicated for postmenopausal osteoporosis, why must the drug not be used during pregnancy?"

Anti-examples (reject internally):
- Storage temperature + adult max dose (two independent instructions).
- "Why is it important to use this drug for bacterial infections, and what is the active ingredient?" (double ask).
- Items answerable from only one section.

CONTEXT A (section: "{title_a}" / label: {label_a}):
{ctx_a}

CONTEXT B (section: "{title_b}" / label: {label_b}):
{ctx_b}

\end{promptbox}

\paragraph{Filtering prompts.} Prompt template for QA correctness filtering of factual questions.

\begin{promptbox}
You are an expert fact-checker evaluating the correctness of an answer.

Your task:
Determine whether the given ANSWER correctly answers the QUESTION based **only** on the information in the CONTEXT.
Ignore any external knowledge. Only rely on the CONTEXT for verification.

Follow these steps internally (do not include your reasoning in the output):
1. Read the QUESTION carefully.
2. Locate relevant parts of the CONTEXT that relate to the QUESTION.
3. Compare the ANSWER to those parts of the CONTEXT.
4. Decide if the ANSWER is fully supported (True) or not supported (False).
5. If the ANSWER is True, extract the exact sentence(s) from the CONTEXT that directly support it.
6. If False, set supporting_evidence to an empty string.

\end{promptbox}

Prompt template for QA correctness filtering of multi-hop questions.

\begin{promptbox}
You are an expert fact-checker evaluating a multi-hop QA item that provides TWO separate context snippets (A and B).

Your task:
Decide whether the QUESTION and ANSWER are (i) supported by BOTH sections and (ii) truly requires BOTH sections (neither A nor B alone is sufficient). Use ONLY the provided contexts.

Follow these steps internally (do not include your reasoning in the output):
1) Read the QUESTION carefully.
2) Examine CONTEXT A and decide if it contains evidence supporting the ANSWER and relevant to the QUESTION.
3) Examine CONTEXT B and decide if it contains evidence supporting the ANSWER and relevant to the QUESTION.
4) Decide whether BOTH sections are needed to answer the QUESTION (i.e., one section alone is insufficient or incomplete).
5) If supported_by_A or supported_by_B is True, extract the minimal supporting sentence(s) from that section.
6) If any field would be False, leave its evidence string empty.

\end{promptbox}

Prompt template for QA correctness filtering of refusal questions.

\begin{promptbox}
You are an expert fact-checker evaluating whether a refusal-style answer is appropriate.

Your task:
Determine whether the provided ANSWER (which indicates "no answer available") 
is correct based **only** on the information in the CONTEXT.
Ignore any external knowledge.

Follow these steps internally (do not include your reasoning in the output):
1. Read the QUESTION carefully.
2. Check if the CONTEXT contains any information that directly answers the QUESTION.
3. If the CONTEXT does not contain any answer, then the refusal ("not found") is correct (True).
4. If the CONTEXT actually contains an answer, then the refusal is incorrect (False).
5. If incorrect, extract the exact sentence(s) from the CONTEXT that show the answer.
6. If correct, leave supporting_evidence as an empty string.

\end{promptbox}

Prompt template for question relevance filtering of all questions.

\begin{promptbox}
Classify the given QUESTION into one of the allowed FDA drug label topics or "None / Irrelevant" if it is not related to regulatory labeling content.

Follow these steps internally (do not include your reasoning in the output):
1. Read the QUESTION carefully.
2. Determine if it concerns factual, regulatory-relevant information that appears in FDA drug labels.
3. If relevant, assign exactly one of the allowed categories.
4. If not relevant, assign "None / Irrelevant".
5. Output a structured JSON object only (no extra text).

Allowed topics (choose exactly one):
1) Indications & Usage
2) Dosage & Administration
3) Dosage Forms & Strengths & Formulation
4) Contraindications
5) Safety & Serious Risks
6) Adverse Events
7) Drug Interactions
8) Use in Specific Populations
9) Pharmacokinetics
10) None / Irrelevant

\end{promptbox}

Prompt template for question quality filtering of factual questions.

\begin{promptbox}
A BAD factual question has one or more of: 
- "answer_leakage": the question already states the key information it is asking for 
- "context_mismatch": the question mentions an entity, age range, dose, or condition not supported by the CONTEXT 
- "unanswerable_from_context": the requested information is simply not present in the CONTEXT 
- "low_utility": technically answerable but extremely vague or trivial as a benchmark item 
- "other": any other serious flaw that makes it unsuitable

\end{promptbox}

Prompt template for question quality filtering of multi-hop questions.

\begin{promptbox}
A BAD multi-hop question has one or more of: 
- "one_sided_answerable": The answer can be derived entirely from Context A alone OR Context B alone, so the question is not truly multi-hop. 
- "hop_mismatch": The question forces an artificial, invalid, or nonsensical connection between Context A and B (e.g., linking unrelated concepts, doses, adverse events, or PK mechanisms across contexts). 
- "answer_leak_from_context": The question copies wording, entities, or details from one context in a way that incorrectly shapes the question (e.g., introduces unsupported conditions, imports irrelevant entities, or fabricates a scenario based on leaked text). 
- "unanswerable_from_context": The requested information does not appear in the contexts; the question cannot be answered from A+B. 
- "other": The question is trivial, vague, or provides negligible benchmarking value.

\end{promptbox}
\clearpage
\section{Experimental details}
\label{app:experimental}

\subsection{Experimental prompts}

\paragraph{Inference prompts.} Prompt template for closed-book setting of QA inference.

\begin{promptbox}
    You are assisting FDA reviewers by answering questions about a single drug label.

General rules:
- Answer as concisely as possible (1-3 sentences).
- Do NOT invent facts that are not supported by the label.

You do NOT have access to the drug label text. Use only your existing knowledge.
    
QUESTION:
{question}

ANSWER:
\end{promptbox}

Prompt template for open full label setting of QA inference.

\begin{promptbox}
    You are assisting FDA reviewers by answering questions about a single drug label.

General rules:
- Answer as concisely as possible (1-3 sentences).
- Do NOT invent facts that are not supported by the label.

The label text below is divided into passages. Each passage is preceded by a marker
of the form: ||PASSAGE_XXXX|| where XXXX is a zero-padded integer (the passage id).

When answering:
- Use ONLY information from the label text.
- After your answer, you MUST list the passage ids that best support your answer.
- If multiple passages are relevant, include all of them.
- If the answer truly cannot be determined from the label, reply exactly with: NOT_ANSWERABLE
- If the answer is NOT_ANSWERABLE, use CITED_PASSAGES: [].
\end{promptbox}

\begin{promptbox}
Output format (exactly):
1. First line: the answer in natural language (or NOT_ANSWERABLE).
2. Second line: CITED_PASSAGES: [PASSAGE_XXXX, PASSAGE_YYYY, ...]

LABEL TEXT:
{label_text}

QUESTION:
{question}
\end{promptbox}

Prompt template for open oracle passages setting of QA inference.

\begin{promptbox}
You are assisting FDA reviewers by answering questions about a single drug label.

General rules:
- Answer as concisely as possible (1-3 sentences).
- Do NOT invent facts that are not supported by the label.

You are given a set of passages extracted from the FDA-approved label
for this drug. Use ONLY these passages to answer.

PASSAGES:
{context_block}

QUESTION:
{question}

ANSWER:
\end{promptbox}

\paragraph{Evaluation prompts.} Prompt template for evaluation of inference answer against gold standard.

\begin{promptbox}
    You are grading answers to questions about FDA-approved drug labels.

You will see:
- a QUESTION about a drug's label,
- a GOLD TARGET: the reference answer derived from the label, and
- a PREDICTED ANSWER: the model's response.

Your job is to decide whether the PREDICTED ANSWER is:
- CORRECT
- INCORRECT
- NOT_ATTEMPTED

and then output a single letter:
- A for CORRECT
- B for INCORRECT
- C for NOT_ATTEMPTED
\end{promptbox}

\begin{promptbox}
## Core grading rules

Treat the GOLD TARGET as the reference truth for what should be said.

A PREDICTED ANSWER is CORRECT if:
- It contains all clinically important information in the GOLD TARGET that is relevant to the QUESTION.
- It does NOT contain any statements that contradict the GOLD TARGET.
- It does NOT introduce specific clinical recommendations that are unsupported or clearly wrong.
- Wording differences are fine (paraphrases, reordered points, different sentences) as long as the meaning matches.

A PREDICTED ANSWER is INCORRECT if:
- It contradicts the GOLD TARGET (different dose, frequency, indication, population, contraindication, etc.), OR
- It introduces specific clinical facts (dose, schedule, indication, contraindication, population, lab threshold, etc.) that are not supported by the GOLD TARGET and would influence clinical use, OR
- It omits one or more MAJOR clinically important elements required by the QUESTION (for example, missing a key dose adjustment, missing a required contraindication), OR
- The GOLD TARGET is a refusal/non-answer (e.g. "Information not found in label") but the model still gives a specific clinical recommendation instead of refusing.

A PREDICTED ANSWER is NOT_ATTEMPTED if:
- It clearly does NOT provide the required information from the GOLD TARGET (e.g. "I don't know", "I cannot answer from the label"), AND
- It does NOT invent or contradict clinical facts in the GOLD TARGET.

## Domain-specific guidance

1. Doses, frequencies, durations, and numeric thresholds
   - If the GOLD TARGET gives a specific dose, frequency, duration, or lab threshold, the PREDICTED ANSWER must have the same key numbers to be CORRECT.
   - Small formatting changes (e.g., "5 mg once daily" vs. "once daily 5 mg") are fine.
   - If the predicted numbers differ in a way that would change dosing or eligibility, grade as INCORRECT.
   - Vague statements like "take as directed on the label" are usually NOT_ATTEMPTED unless the GOLD TARGET itself is vague.
\end{promptbox}

\begin{promptbox}
2. Indications and populations
   - If the GOLD TARGET specifies indications or special populations (e.g., "patients with eGFR < 45 mL/min/1.73 m2", "pediatric patients 6-17 years"), leaving out a major constraint or population can make the answer INCORRECT.
   - Minor wording differences (e.g., "patients with moderate to severe renal impairment" when the GOLD TARGET explicitly defines that range) can still be CORRECT if they preserve the same meaning.

3. Contraindications and warnings
   - If the QUESTION asks about contraindications or major warnings, missing a key contraindication or warning from the GOLD TARGET should be graded as INCORRECT, not NOT_ATTEMPTED.
   - Adding a serious new contraindication or warning that is not in the GOLD TARGET is INCORRECT, even if it sounds medically plausible.

## Examples

### Example 1 (dose and population)

Question:
"What is the recommended saxagliptin dose for patients with eGFR < 45 mL/min/1.73 m2?"

Gold target:
"The recommended dosage of saxagliptin tablets is 2.5 mg orally once daily for patients with eGFR < 45 mL/min/1.73 m2, including those with moderate or severe renal impairment or ESRD."

Predicted answer 1:
"Give 2.5 mg saxagliptin once daily in patients with eGFR below 45. This includes patients with moderate or severe renal impairment and ESRD."

- A: contains the key dose and population, no contradictions.

Predicted answer 2:
"Use the standard 5 mg once daily dose regardless of renal function."

- B: contradicts the GOLD TARGET on dose and population.

Predicted answer 3:
"I'm not sure what dose to use in patients with reduced kidney function based on this label."

- C: does not provide the needed information.
\end{promptbox}

\begin{promptbox}
### Example 2 (contraindication)

Question:
"In which patients are potassium citrate tablets contraindicated?"

Gold target:
"Potassium citrate extended-release tablets are contraindicated in patients with hyperkalemia or conditions predisposing them to hyperkalemia, patients with GI obstruction or delayed gastric emptying, patients with peptic ulcer disease, patients with active urinary tract infection with certain stones, and patients with renal insufficiency."

Predicted answer 1:
"They are contraindicated in patients with hyperkalemia or at risk of hyperkalemia, GI obstruction or delayed gastric emptying, peptic ulcer disease, active urinary tract infection with certain stones, and renal insufficiency."

- A: contains the correct information.

Predicted answer 2:
"They are contraindicated only in patients with a history of allergies to potassium."

- B: omits almost all key contraindications and adds an unsupported one.

Predicted answer 3:
"The label does not clearly specify in which patients they are contraindicated."

- C: fails to use the GOLD TARGET, but does not contradict it

## Final instruction

Now you will grade a new example.

You will be given:

Question: {question}
Gold target: {target}
Predicted answer: {predicted_answer}

Grade the predicted answer of this new question as one of:
A: CORRECT
B: INCORRECT
C: NOT_ATTEMPTED
\end{promptbox}

\begin{promptbox}
Respond in the following format, on a single line:

LETTER: short reason

Where LETTER is exactly one of A, B, or C, and "short reason" is 1-2 sentences explaining your choice.
Do not include any other text.
\end{promptbox}

\section{AI Use Declaration}
During the preparation of this manuscript, the authors used ChatGPT for assistance purely with the editing of the paper. \clearpage
\onecolumn
\section{Expert annotator rubrics}
\label{app:rubric}

\scriptsize
\setlength{\tabcolsep}{3pt}
\renewcommand{\arraystretch}{1.15}

\begin{longtable}{
    r
    >{\RaggedRight\arraybackslash}p{0.38\textwidth}
    >{\RaggedRight\arraybackslash}p{0.11\textwidth}
    >{\RaggedRight\arraybackslash}p{0.17\textwidth}
    >{\RaggedRight\arraybackslash}p{0.25\textwidth}
}
\caption{Expert relevance annotations for benchmark questions. Question column is provided to the expert annotator, and they fill out the relevance (relevant / partially relevant / irrelevant), category, and provide any reasons.}
\label{tab:expert_relevance_annotations} \\

\toprule
\textbf{ID} & \textbf{Question} & \textbf{Relevance} & \textbf{Category} & \textbf{Reason} \\
\midrule
\endfirsthead

\multicolumn{5}{l}{\textit{Table \thetable{} continued from previous page.}} \\
\toprule
\textbf{ID} & \textbf{Question} & \textbf{Relevance} & \textbf{Category} & \textbf{Reason} \\
\midrule
\endhead

\midrule
\multicolumn{5}{r}{\textit{Continued on next page.}} \\
\endfoot

\bottomrule
\endlastfoot
1 & What is the recommended dosage of Plavix for patients with vascular disease, and what was the outcome of its use in the CHARISMA trial? & partially relevant & Dosage \& Administration & Half of question is regarding dosage instruction, other half is about trial outcome \\
2 & In patients taking metformin, how might the risk of vitamin B12 deficiency influence the management of overdose cases? & partially relevant & Safety \& Serious Risks & Question related to management of overdoses is relevant and b12 deficiency is a relevant warning, but they are not related. \\
3 & What is the indication for Felodipine and how can gingival hyperplasia be managed in patients receiving this treatment? & relevant & Indications \& Usage & Asks directly about indication then about management of AE \\
4 & What is the pharmacokinetic equivalence of carvedilol phosphate extended-release capsules compared to immediate-release carvedilol tablets in treating heart failure? & relevant & Pharmacokinetics & Directly asks about PK of drug. \\
5 & Why is metaxalone contraindicated in patients with severe hepatic impairment despite the unknown effects of hepatic impairment on its pharmacokinetics? & relevant & Contraindications & Directly asks about contraindication \\
6 & What is the indication for Carbatrol in adults, and what is the lowest known lethal dose for this population? & relevant & Indications \& Usage & First half of question is highly relevant to indication. Lethal dose/acute toxicity is a safety concern for this product \\
7 & Why should Gleostine be used with caution in patients with a baseline Forced Vital Capacity below 70\% when considering the risk of secondary malignancies? & partially relevant & Safety \& Serious Risks & Directly asking about use in patients with low FVC, but FVC is pulmonary toxicity. Not related to secondary malignancy \\
8 & What is the absolute bioavailability of JESDUVROQ (daprodustat)? & relevant & Pharmacokinetics & Clearly asks about PK of drug \\
9 & What is the INR cutoff for beginning Desipramine Hydrochloride administration in patients experiencing severe hepatic impairment? & irrelevant & None / Irrelevant & Information for INR or hepatic impairment not in labeling \\
10 & What is the recommended frequency for INR assessments in hemodialysis patients on topcare dual action complete therapy? & irrelevant & None / Irrelevant & Question about monitoring for INR. This is not correct for this product (not in label) \\
11 & What weight loss outcomes can be expected in patients treated with ORLISTAT compared to placebo, considering the common gastrointestinal adverse events associated with its use? & irrelevant & None / Irrelevant & Overall question is asking about an outcome \\
12 & What is the pregnancy prevention rate for women aged 18 to 35 taking LAYOLIS Fe, and what percentage of women discontinued the trial due to adverse reactions? & irrelevant & None / Irrelevant & About treatment outcome and clinical trial \\
13 & What are the available strengths of Minocycline hydrochloride extended-release tablets for acne treatment despite the unknown mechanism of action? & partially relevant & Dosage Forms \& Strengths \& Formulation & First half is about strengths and second half is regarding MOA. \\
14 & What monitoring strategies should be employed for QTc interval in hemodialysis patients treated with Clindamycin Palmitate Hydrochloride? & irrelevant & None / Irrelevant & QTC not warning for clindamycin \\
15 & What is the recommended dosing range for alprazolam in adults with generalized anxiety disorder given its pharmacokinetic profile? & relevant & Dosage \& Administration & Asks about dosing. But, dosing may or may not be related to PK for this product \\
16 & What dosage forms of ORSERDU are available for geriatric patients aged 65 and older? & partially relevant & Dosage Forms \& Strengths \& Formulation & Question is broadly asking about different dosage forms. Which may require info from multiple labels \\
17 & What is the expected pregnancy rate for typical users of Lyleq compared to the pharmacokinetics of norethindrone absorption? & irrelevant & None / Irrelevant & Overall question is asking about an outcome. Incorrectly mentions PK \\
18 & What precautions should be taken when initiating captopril in patients on diuretics to avoid hypotension? & relevant & Safety \& Serious Risks & About hypotension which is warning in label \\
19 & What serious adverse event has been reported in patients treated with acamprosate calcium that is not mentioned in the labeling, and how does food affect its bioavailability? & irrelevant & None / Irrelevant & Directly asks for info not in label \\
20 & What should be done regarding routine contraception after using levonorgestrel tablets, 0.75 mg, considering the higher pregnancy rates observed in Chinese women? & irrelevant & None / Irrelevant & About routine contraception not related to specific product. Also asking about an outcome \\
21 & What is the recommended use of Pindolol in patients with hypertension when considering concomitant therapy with thiazide-type diuretics? & relevant & Indications \& Usage & Directly asking about indication per labeling \\
22 & How does smoking affect the pharmacokinetics of erlotinib in patients? & relevant & Drug Interactions & About smoking effect on PK \\
23 & What is the lowest known lethal dose of Tegretol for adults, and how does this inform dosing considerations in patients with a history of substance use disorder? & partially relevant & Safety \& Serious Risks & First half of question is about safety, second half is irrelevant to this drug product \\
24 & What is the peak serum concentration time for enalapril after oral administration of enalapril maleate tablets? & relevant & Pharmacokinetics & Directly asks about PK of drug \\
25 & What gastrointestinal adverse reactions are associated with clindamycin use? & relevant & Adverse Events & Directly asks about AEs \\
26 & What is the volume of distribution for ethionamide in healthy volunteers treated with Trecator for tuberculosis? & relevant & Pharmacokinetics & Relevant because asking about PK, but in the question has grouped healthy volunteers together with tb patients. \\
27 & What lifestyle modifications should be attempted before starting LOPID therapy in patients with lipid abnormalities, considering its contraindications with certain statins? & irrelevant & None / Irrelevant & Question is about lifestyle modifications, not about contraindication. Lifestyle modifications not in label \\
28 & What is the recommended frequency for INR monitoring in hemodialysis patients administered equaline loperamide hydrochloride? & irrelevant & None / Irrelevant & INR not relevant for this product. Also about monitoring \\
29 & What are the available strengths of Dimethyl Fumarate Delayed-Release Capsules? & relevant & Dosage Forms \& Strengths \& Formulation & Directly asking about strength \\
30 & What are the indications for Carisoprodol and Aspirin Tablets? & relevant & Indications \& Usage & Asks about indication \\
31 & What is the recommended INR monitoring schedule for patients on hemodialysis who are administered Allegra Allergy? & irrelevant & None / Irrelevant & INR not relevant for this product. Also about monitoring \\
32 & What are the signs and symptoms of acute overdosage with phendimetrazine tartrate? & relevant & Safety \& Serious Risks & Directly asks about overdosage which is noted in label \\
33 & What role does Anti-Diarrheal, Caseys 4good play in the treatment of liver function test issues in pediatric patients? & partially relevant & Safety \& Serious Risks & Combining label information about liver impairment precaution with other information \\
34 & What effect does estrogen administration have on thyroid-binding globulin (TBG) concentrations in women taking Angeliq? & relevant & Safety \& Serious Risks & About labeled warning \\
35 & What should patients do if they are taking or plan to take other medications while on bupropion hydrochloride extended-release tablets (XL)? & irrelevant & None / Irrelevant & Not related to available categories \\
36 & Why should NUZYRA not be prescribed in patients without a confirmed bacterial infection, considering the risk of Clostridioides difficile-associated diarrhea? & partially relevant & Safety \& Serious Risks & First half of question is regarding warning about drug-resistant bacteria. Second half incorrectly links to C diff. \\
37 & What is the effect of fluvoxamine on the AUC and Cmax of ramelteon? & relevant & Contraindications & Asking about drug interaction, but is a labeled contraindication \\
38 & What is the strength of VELSIPITY tablets and what respiratory monitoring is recommended for patients with asthma during treatment? & partially relevant & Dosage Forms \& Strengths \& Formulation & Strength relevant, but monitoring in asthma is not \\
39 & Why should Acid Reducer Complete not be used in patients with unexplained weight loss when it contains famotidine? & partially relevant & Safety \& Serious Risks & Unexplained weight loss is a warning per label, reasons why and relation to famotidine are not. \\
40 & What is the dosage form of aripiprazole and what metabolic changes should be monitored in patients receiving it? & relevant & Dosage Forms \& Strengths \& Formulation & About dosage forms. Second half is about metabolic changes precaution \\
41 & What role does HealthMart play in addressing inflammatory markers in G6PD-deficient patients? & irrelevant & None / Irrelevant & Not about a drug? \\
42 & Why must Raloxifene hydrochloride tablets be stored at controlled room temperature despite being contraindicated in pregnancy? & partially relevant & Contraindications & Storage is irrelevant, but contraindication in pregnancy is from label. Incorrectly connected storage and contraindication \\
43 & What precautions should be taken when prescribing SYMBRAVO to patients who are also using rizatriptan for migraine treatment? & irrelevant & None / Irrelevant & Unclear question. Rizatriptan is component of Symbravo \\
44 & Can EQUETRO be used with other antiepileptic drugs (AEDs)? & relevant & Dosage \& Administration & Information is mentioned in dosage section. Elaborated on in drug interactions section \\
45 & What is the primary indication for EPZICOM in the treatment of HIV-1 infection? & relevant & Indications \& Usage & Relevant, but indication is answered in the question \\
46 & What monitoring is necessary for patients on calcitriol to prevent complications from uncontrolled calcium intake? & partially relevant & Safety \& Serious Risks & Hypercalcemia is a risk with product. Not necessarily caused by calcium intake \\
47 & What is the QTc interval cutoff for commencing treatment with Gencare-Allergy Relief Loratadine Tablets, 10 mg in pregnant individuals? & irrelevant & None / Irrelevant & QTC not related to loratidine \\
48 & What is the recommended approach for adjusting CareOne Acid Reducer Complete dosage when inflammatory markers surpass normal levels? & irrelevant & None / Irrelevant & Not relevant to drug product \\
49 & What side effects might I experience after taking Plan B One-Step if I used it more than 72 hours after unprotected sex? & irrelevant & None / Irrelevant & >72 hours is outside of labeled use time \\
50 & What is the initial recommended dose of RECORLEV for patients without active liver disease, and why is it contraindicated in those with cirrhosis? & relevant & Dosage \& Administration & Relevant to dosage and contraindications \\
\end{longtable}

\end{document}